%% file: main.tex
\documentclass{article}

\usepackage{microtype}
\usepackage{graphicx}
\usepackage{booktabs} 

\usepackage[accepted]{icml2024}

\usepackage{amsmath}
\usepackage{amssymb}
\usepackage{mathtools}
\usepackage{amsthm}

\usepackage[utf8]{inputenc} 
\usepackage{booktabs}       
\usepackage{amsfonts}       
\usepackage{nicefrac}       
\usepackage{microtype}      
\usepackage{xcolor}         

\usepackage{url}

\usepackage{breakurl}
\usepackage[breaklinks]{hyperref}

\usepackage{graphicx}
\usepackage{wrapfig}
\usepackage[capitalize,noabbrev]{cleveref}
\usepackage{subcaption}
\usepackage[bottom]{footmisc}
\usepackage{multirow}
\usepackage{textgreek}
\usepackage{dblfloatfix}
\usepackage{enumitem}

\input{math_commands.tex}

\theoremstyle{plain}

\theoremstyle{definition}

\theoremstyle{remark}

\usepackage[textsize=tiny]{todonotes}

\icmltitlerunning{Memoria: Resolving Fateful Forgetting Problem through Human-Inspired Memory Architecture}

\allowdisplaybreaks

\begin{document}

\twocolumn[
\icmltitle{Memoria: Resolving Fateful Forgetting Problem through \\ Human-Inspired Memory Architecture}




\begin{icmlauthorlist}
\icmlauthor{Sangjun Park}{skku}
\icmlauthor{JinYeong Bak}{skku}
\end{icmlauthorlist}

\icmlaffiliation{skku}{Department of Computer Science and Engineering, Sungkyunkwan University, Suwon, South Korea}

\icmlcorrespondingauthor{JinYeong Bak}{jy.bak@skku.edu}

\icmlkeywords{Machine Learning, ICML}

\vskip 0.3in
]



\printAffiliationsAndNotice{}  

\begin{abstract}
    \input{pages/Abstract}
\end{abstract}

\section{Introduction}
\label{sec:introduction}
\input{pages/Introduction}

\section{Background}
\label{sec:background}
\input{pages/Background}

\section{Memoria}
\label{sec:memoria}
\input{pages/Memoria}

\section{Memoria-Applied Transformers}
\label{sec:memoria-transformers}
\input{pages/Memoria-Transformers}

\section{Experiments}
\label{sec:experiments}
\input{pages/Experiments}

\section{Psychological Memory Effects}
\label{sec:psychological-memory-effects}
\input{pages/Psychological-Memory-Effects}

\section{Conclusion and Future Work}
\label{sec:conclusion}
\input{pages/Conclusion}

\section*{Impact Statement}
\label{sec:impact-statement}
\input{pages/Impact-Statement}

\section*{Acknowledgements}
We would like to thank the anonymous reviewers for their helpful questions and comments.
This project is partially supported by Microsoft Research Asia.
This research was partly supported by the Bio \& Medical Technology Development Program of the National Research Foundation (NRF) funded by the Korean government (MSIT) (NRF-2021M3A9E4080780),  
Institute of Information \& communications Technology Planning \& Evaluation (IITP) grant funded by the Korea government (MSIT) (IITP-2023-2020-0-018, abductive inference framework using omni-data for understanding complex causal relations \& ICT Creative Consilience program and RS-2024-00398115, Research on the reliability and coherence of outcomes produced by Generative AI).

\bibliography{memoria}
\bibliographystyle{icml2024}


\newpage
\appendix
\onecolumn

\section{Hebbian Attributes for Memoria}
\label{sec:hebbian-attributes-of-memoria}
\input{pages/appendix/Hebbian-attributes-of-memoria}

\newpage

\section{Details on Psychological Memory Effects}
\label{sec:psychological-memory-effects-details}
\input{pages/appendix/Details-On-Psychological-Memory-Effects}

\newpage

\section{Ablation Study}
\label{sec:ablation-study}
\input{pages/appendix/Ablation-Study}

\section{Training Details and Additional Results}
\label{sec:additional-experiment}
\input{pages/appendix/Training-Details-And-Additional-Results}

\section{Algorithm \& Computational Complexity}
\label{sec:algorithm-and-computational-complexity}
\input{pages/appendix/Algorithm-And-Computational-Complexity}

\section{RAG Discussions}
\label{sec:rag-discussions}
\input{pages/appendix/RAG-Discussions}

\newpage

\section{Structure of Memoria-Applied Transformers}
\label{sec:memoria-applied-transformers-structure}
\input{pages/appendix/Memoria-applied-transformers}

\newpage

\section{Visualization of Memoria}
\label{sec:visualization-of-memoria}
\input{pages/appendix/Visualization-Of-Memoria}

\end{document}

%% file: math_commands.tex
\usepackage{amsmath,amsfonts,bm}

\def\eqref#1{equation~\ref{#1}}

\def\1{\bm{1}}

\DeclareMathAlphabet{\mathsfit}{\encodingdefault}{\sfdefault}{m}{sl}
\SetMathAlphabet{\mathsfit}{bold}{\encodingdefault}{\sfdefault}{bx}{n}

\DeclareMathOperator*{\argmax}{arg\,max}

%% file: pages/Abstract.tex
Making neural networks remember over the long term has been a longstanding issue.
Although several external memory techniques have been introduced, most focus on retaining recent information in the short term.
Regardless of its importance, information tends to be fatefully forgotten over time.
We present Memoria, a memory system for artificial neural networks, drawing inspiration from humans and applying various neuroscientific and psychological theories.
The experimental results prove the effectiveness of Memoria in the diverse tasks of sorting, language modeling, and classification, surpassing conventional techniques.
Engram analysis reveals that Memoria exhibits the primacy, recency, and temporal contiguity effects which are characteristics of human memory. 

%% file: pages/Introduction.tex
Humans possess an incredible ability to retain memories for long periods.
Humans extract the gist from the flood of data, retrieve relevant information, and gradually forget useless and unemployed memories.
Efforts to endow neural networks with human-like long-term memory have been ongoing.
Although Transformers \citep{transformer} have shown excellent performance in a variety of tasks \citep{bert, gpt1, gpt, bart}, they also struggle with long sequences due to the nature of processing entire input tokens simultaneously.
To mitigate this limitation, external memory methodologies have been studied.
Nevertheless, unlike humans, most existing methods prioritize the preservation of new information over old memories and operate with fixed capacities. Consequently, this inevitably leads to the removal or dilution of old memories. We termed this problem \textbf{\emph{Fateful Forgetting}}.

Introducing a dynamic memory capacity and employing a policy that prioritizes crucial information for the future can resolve the issue of fateful forgetting. However, realizing this requires addressing various derivative problems. Firstly, it is necessary to distinguish which information is considered important \emph{\textbf{(Long-term Importance)}}.
Predicting long-term importance at the initial acquisition is challenging, as the determination of whether it will be useful in the future depends on future usage, making it difficult to foresee.
Furthermore, since we cannot store infinite amounts of information, forgetting is essential. This mechanism should not merely erase old information, but rather selectively preserve and forget based on the long-term importance of the information \emph{\textbf{(Selective Preservation)}}.
Furthermore, while recent memories inherently preserve a certain degree of relevance to the context, long-term memories are not the case.
Because long-term memories are temporally distant from the current situation, the content of the retrieved long-term memory should be relevant to the current situation.
Ultimately, it is crucial that old memories are selectively activated based on the current context \emph{\textbf{(Cue-based Activation)}}. This issue encompasses the challenge of how to search associated memories in long-term storage \emph{\textbf{(Memory Searching)}}.

Fortunately, all these issues have been long-standing challenges faced by the memory system of living organisms. Humans possess a highly sophisticated memory system that not only retains recent information but also has the capability to remember important events throughout their lives \citep{multistore-model, levels-of-processing, adaptive-memory, displacement,trace-decay-brown, interference}.
Recent advancements in the fields of AI and neuroscience have brought attention to the significance of interdisciplinary research between these two domains \citep{neuro-inspired-ai, brain-inspired-replay}.
Particularly, in the realm of memory systems, humans provide nearly ideal solutions, prompting endeavors to apply the insights from human memory systems to artificial neural networks \citep{memo, machine-memory-system}.
Following this trend, we approach the issue of fateful forgetting by integrating the neuroscientific evidence and the theoretical models of human memory.
Memoria provides an innovative solution to fateful forgetting, opening the pathway to selective and permanent memorization for neural networks.

\begin{figure*}[t]
    \centerline{\includegraphics[width=\textwidth]{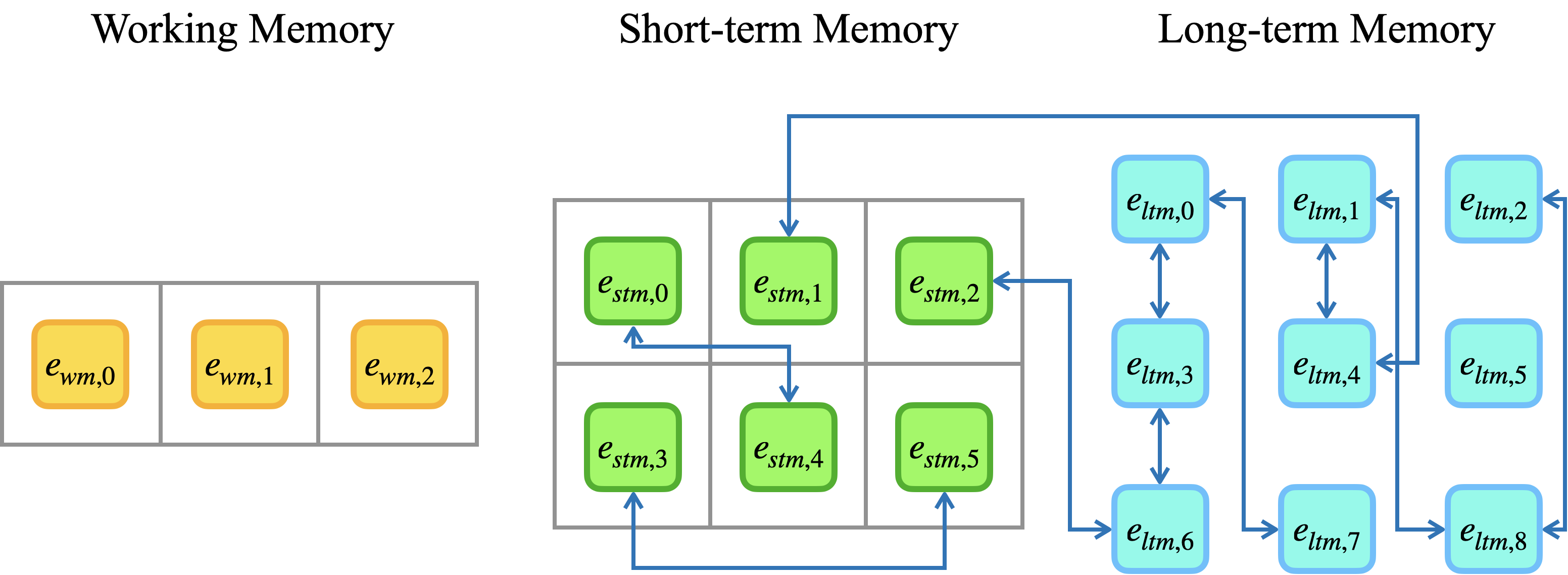}}
    \caption{Working memory retains the most recent information. Short-term memory also holds a fixed number of recent engrams. The number of engrams in long-term memory is not predetermined. The arrows in the diagram represent the connections between each engram. Each connection is directed and weighted. Those weighted edges are used for retrieval.}
    \label{fig:memoria-structure}
\end{figure*}

\textbf{Contributions}

1. We designed Memoria as an external memory framework for neural networks, incorporating various theories of human memory\footnote{The implementation of Memoria and all experimental code are publicly available at \url{https://github.com/cosmoquester/memoria}}.
We provide evidence that Memoria successfully addresses fateful forgetting via extensive analysis.

2. We effectively integrated Memoria into GPT, BERT, and RoBERTa representing its superior performance in comparison to the traditional external memory methodologies across the sorting, language modeling, and text classification tasks.

3. We discovered the similarity of long-term memory between Memoria and humans by showing that Memoria closely reproduces the three well-known effects of human memory: primacy, recency, and temporal contiguity effects.

%% file: pages/Background.tex
\textbf{Memory of Neural Networks }
Recurrent Neural Networks \citep{rnn, lstm, gru} were introduced to process sequential data. Memory Augmented Neural Networks (MANNs) emerged to perform complex memory operations beyond simple sequential processing. Neural Turing Machines (NTMs) \citep{ntm} have a storage system that can be accessed using an attention mechanism. NTMs were further developed into DNC \citep{dnc}, Sparse DNC \citep{sparse-dnc}, D-NTM \citep{d-ntm}, TARDIS \citep{tardis}, and GCL \citep{gcl}. 
After the success of Transformer, research has focused on Transformer's limited context length.

Two major approaches have been proposed to address this limitation.
The first approach involves computational optimization of architectures such as Longformer \citep{longformer}, BigBird \citep{bigbird}, and Reformer \citep{reformer}.
However, the models still process only a restricted size of inputs, even though they handle longer lengths with the same amount of resources.
The second approach involves leveraging external memory storage, exemplified by models such as Transformer-XL \citep{transfoxl}, Compressive Transformer \citep{compformer}, $\infty$-Transformer \citep{infformer}, Memory Transformer \citep{memory-transformer}, Recurrent Memory Transformer \citep{rmt}, and Memorizing Transformers \citep{memorizing}.
These models split inputs into multiple segments and incorporate them to better maintain long-term dependencies in sequential data.
They have a simpler structure compared to traditional MANNs and utilize memory focused on recent information.
Therefore, in most cases among them, they are not immune to the issue of fateful forgetting.
Memoria also follows the second approach, but overcomes fateful forgetting by imitating the human mind.

Memoria categorizes memories into three levels according to the Multi-Store model \citep{multistore-model}, using the term working memory instead of sensory memory. Memoria relies on two mechanisms of forgetting. 
Firstly, for forgetting in short-term memory, we applied the displacement mechanism \citep{displacement}, which replaces old information with new information.
Secondly, for forgetting in both short-term and long-term memory, we incorporated the concept of trace decay theory \citep{trace-decay-brown, trace-decay-peter}, which suggests that memories gradually fade away if they are not actively recalled. This strategy assists Memoria to preserve useful memories.

\textbf{Long-term Importance } According to the Multi-Store Model \citep{multistore-model}, memories are better preserved and consolidated through repetitive rehearsal. As frequently accessed memories are easy to retain for longer duration \citep{role-of-retrieval-practice, retrieval-fast-route}, Memoria prioritizes maintaining the repeatedly recalled memories.
Memoria updates this information at each time step to maintain the long-term importance of each memory.

\textbf{Selective Preservation } The following problem is determining how to selectively preserve only those distinguished memories.
Humans make use of diverse forgetting strategies \citep{trace-decay-brown, trace-decay-peter, interference, displacement}.
Memoria employs decay as a key forgetting mechanism, assigning a predetermined lifespan to each memory and constantly decreasing its lifespan. The only way memories acquire lifespan is through retrieval and utilization.
This design ensures that lifespan is obtained proportionally to the degree of contribution, allowing important memories to persist for a long time.
This reflects the brain's characteristics of preserving memories associated with usefulness and high rewards in the long-term \citep{how-brain-maintain-useful-memory, brain-keep-high-reward-memory}.

\textbf{Cue-based Activation } The cue-based activation and memory searching problems are related to memory retrieval. SAM \citep{sam-origin, sam, sam-retro}
is a standard for subsequent memory models \citep{compu-model-memory-search}.
The concept of global matching in SAM is widely accepted, where the associative weights between the current context and memory are employed in retrieval.
Similarly, in Memoria, a working memory always represents the most recent memory, resolving the cue-based activation problem by leveraging its distance from retrieval candidates.

\textbf{Memory Searching } In searching memory \citep{storage-and-retrieval, search-process-recognition, search-decision-process}, we adopted the concept of global searching, which is a key feature of SAM \citep{global-neural-similarity}. SAM not only reflects the association between context and memory but also considers the mutual association among memory pieces. SAM initially retrieves memories from long-term memory using the association with short-term memory. Once a new memory is recalled, the memory is used to iteratively recall further memories by leveraging the association with previously retrieved memories. This iterative process increases association among memories recalled together, facilitating their easy recall in subsequent retrievals. Memoria, similarly, resolves the memory searching problem by interconnecting individual memory pieces and employs a mechanism to search for next memory based on the recalled.

\textbf{Hebbian Theory } An engram serves as the fundamental unit of memory in neuroscience, with long-term potentiation (LTP) of synaptic strength operating as a central mechanism in engram formation \citep{what-is-memory-engram}. Hebbian theory \citep{Hebb1949} is a prominent neural plasticity theory that postulates how connections between two neurons change. LTP is one of the key concepts of Hebbian theory, suggesting that when two neurons are repeatedly activated together, their interconnection is strengthened. This phenomenon is commonly referred to as the ``Fire together, wire together'' principle.
In recent years, there has been growing interest in applying Hebbian learning to deep learning \citep{hebb-nn, hebb-without-feedback}. Some studies \citep{fast-parametric, h-mem, self-attentive-asso, hopfield-all-you-need} modeled associative memory using neural networks. Hebbian learning rule \citep{hebbian-learning-rule, hebb-competitive}, a mathematical formulation of Hebbian learning, specifically articulates the process of Hebbian learning.
Memoria also treats engrams as the minimal unit of memory, and the weight changes of engrams are designed to follow Hebb's rule.
Further, \cref{sec:hebbian-attributes-of-memoria} shows that Memoria satisfies all six crucial mathematical properties \citep{math-hebb} for Hebbian learning rule even though it does not take the typical mathematical form of Hebbian learning.

%% file: pages/Memoria.tex
\begin{figure*}[tb!]
    \centerline{\includegraphics[width=\textwidth]{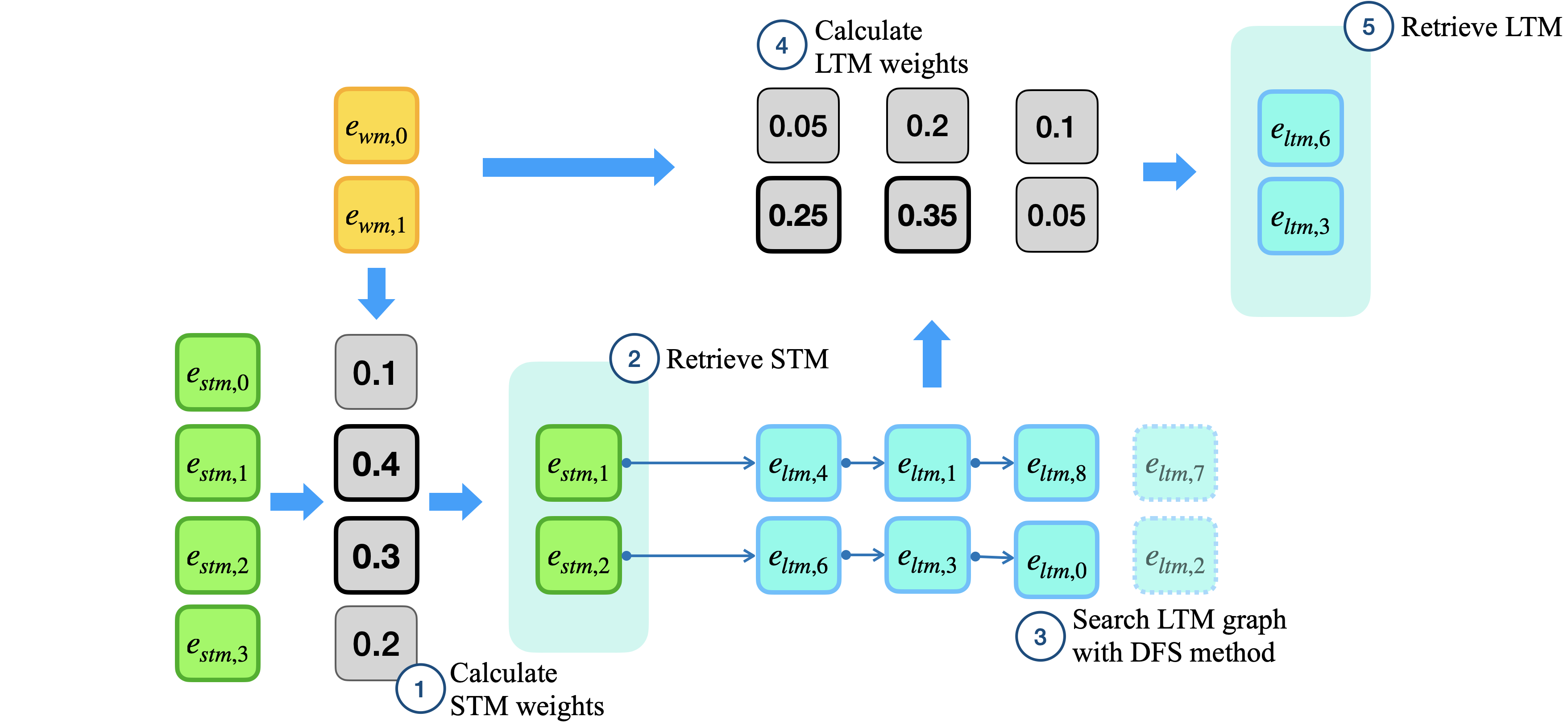}}
    \caption{Retrieval process in Memoria. Memoria utilizes working memory to identify associated engrams in both short-term and long-term memory. The calculated weights in steps 1 and 4 mean the strength of association between the engrams and working memory, with larger values leading to the final selection of the engram. This mechanism deals with the cue-based activation problem by reflecting the association with working memory. Engrams in the gray area represent retrieved engrams.}
    \label{fig:retrieve-process}
\end{figure*}

There are three stages of utilizing Memoria. The first stage is \hyperref[subsec:retrieve]{\textit{retrieve}} stage, in which it uses working memory as a cue to retrieve the engrams from short-term memory and long-term memory. The second stage is \hyperref[subsec:exploit]{\textit{exploit}} stage, where a model uses the retrieved engrams to solve the task. The last stage is \hyperref[subsec:memorize-and-forget]{\textit{memorize \& forget}}. In this stage, all retrieved engrams get more lifespan depending on the usefulness of each engram, and all the engrams lose their lifespan by one.

\subsection{Component}

Memoria consists of three distinct types of memory: working memory, short-term memory, and long-term memory, each composed of engrams.
In \cref{fig:memoria-structure}, we depict the overall structure of the three memory components in Memoria. 

\textbf{Engram} An engram is the smallest unit of memory information, and engrams constitute each memory.
Each engram possesses its own unique lifespan and connection weights.
Memoria continuously preserves engrams in sequences and appropriately manages the lifespans and connection weights of engrams.
In this study, we treated the information part of an engram as an embedding vector (engram $e \in R^d$, where $d$ is the model dimension) for all experiments. However, engrams can theoretically take various forms. When using engrams of different forms, a correlation function between the two engrams should be defined. For instance, one could define a single engram as a text sentence and utilize a correlation function based on edit distance to employ Memoria.

\textbf{Working Memory}
Working memory (WM) is the repository of immediate memory and serves as a reference to access associated engrams from short-term and long-term memory. Working memory adopts a queue structure whose size is fixed by the number of newly created engrams in a single time step. At every time step, working memory is updated.

\textbf{Short-term Memory} Short-term memory (STM), denoted as $M_{stm}$, holds relatively recent information. The engrams in working memory are transferred to short-term memory when new information comes. Short-term memory employs a queue data structure with a configurable and fixed capacity.

\textbf{Long-term Memory} Long-term memory (LTM), denoted as $M_{ltm}$,
has the capacity to store an indefinite number of engrams. The dequeued engrams from short-term memory are transferred to long-term memory.
Long-term memory preserves a wide range of information, spanning from the earliest recollections to recent ones. Therefore, the engrams of long-term memory vary in their creation times and ages.

\textbf{Memory Graph} Engrams in any memory can be linked together, forming a directed weighted graph data structure, where each vertex corresponds to an engram. A directed edge weight $E_{i \rightarrow j}$ indicates the empirical conditional probability of retrieving engram $e_j$ after engram $e_i$ has been retrieved, with $M^{rem}$ representing the set of all retrieved engrams.
This probability is determined by dividing the number of times $e_i$ and $e_j$ were retrieved together ($Count_{i,j}$) by the number of times $e_i$ was retrieved ($Count_{i,i}$).
These weighted edges facilitate finding engrams in long-term memory, with their weights adjusted according to the ``Fire together, wire together'' principle \citep{Hebb1949}.
This graph structure mirrors the association weights between items in SAM \cite{sam-origin, compu-model-memory-search}, playing a pivotal role in addressing the memory search problem.

\begin{equation*}
    \begin{aligned}
        E_{i \rightarrow j} &= P(e_j \in M^{rem} \mid e_i \in M^{rem}) \\
        &= \frac{Count_{i,j}}{Count_{i,i}}
    \end{aligned}
\end{equation*}

\begin{figure*}
    \centerline{\includegraphics[width=\textwidth]{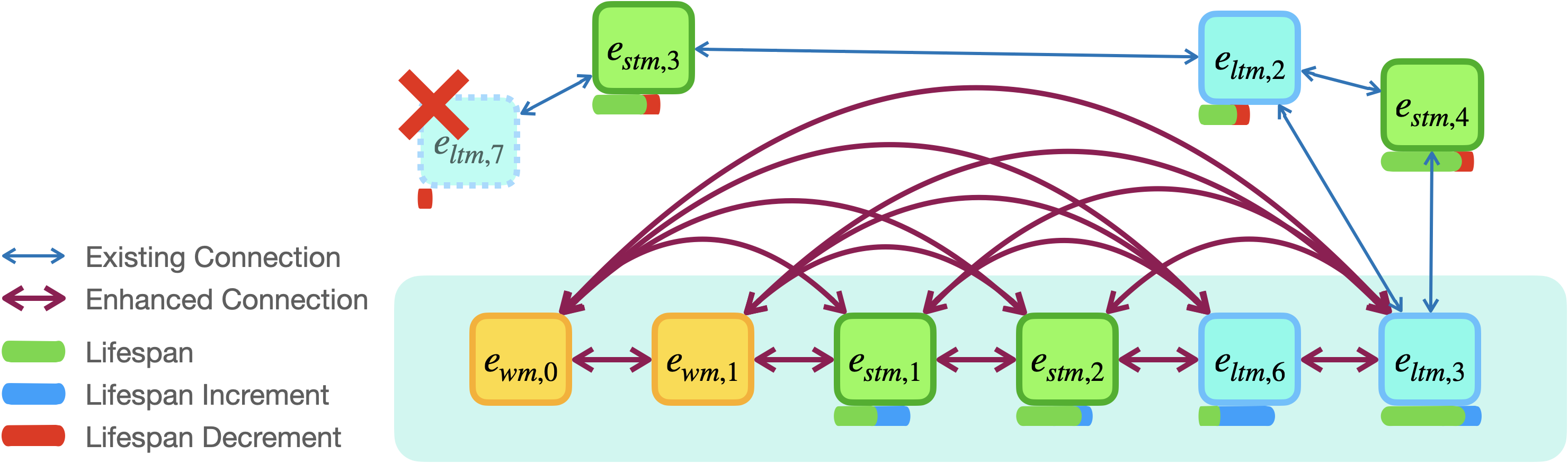}}
    \caption{The connections between working memory and retrieved engrams are strengthened across all pairs. The lifespan of retrieved engrams extends proportionally to their individual contribution, enabling selective preservation through the differential allocation of lifespans.
    Engrams having lost all their lifespan, exemplified by $e_{ltm,7}$, are eliminated permanently.
    }
    \label{fig:memorize-and-forget}
\end{figure*}

\subsection{Retrieve}
\label{subsec:retrieve}

In this stage, retrieval of adequate engrams is conducted by exploring short-term and long-term memory based on working memory.
\cref{fig:retrieve-process} shows entire retrieving process.

\begin{enumerate}[itemsep=1ex,topsep=0ex,partopsep=0ex,parsep=0ex]
\item Replace working memory $M_{wm}$ with new engrams. All the engrams in working memory will have the same initial lifespan. $N_{wm}$ means the number of working memory engrams.

\begin{equation*}
    M_{wm} = \{ e_{wm,1}, e_{wm,2}, \dots, e_{wm, N_{wm}} \}
\end{equation*}

\item By utilizing the correlation function $f_c$, calculate the correlation weight $C_{stm}$ for each $e_{stm,i}$ within short-term memory $M_{stm}$ by averaging all the correlation weights for the engram. The distance function $f_d$ used is L2 distance. Here, $i$ represents the index of $M_{stm}$ and $j$ represents the index of $M_{wm}$.

\begin{equation*}
\begin{aligned}
    f_c(e_i, e_j) &= \mathrm{exp}(-(f_d(e_i, e_j))^2) \\
    C_{stm,i} &= \frac{1}{N_{wm}} \sum_{j = 1}^{N_{wm}} f_c(e_{stm,i}, e_{wm,j})
\end{aligned}
\end{equation*}

\item Select only the top $N_{stm}^{rem}$ number of engrams with $C_{stm}$ values to retrieve. Denote the selected engrams as $M_{stm}^{rem}$.

\item For each $e_i \in M_{stm}^{rem}$, select an engram in $M_{ltm}$ having highest edge weight from $e_i$. Denote the selected engrams as $M_{ltm}^{init}$.

\begin{gather*}
    M_{ltm}^{init} = \argmax_{e_j \in M_{ltm}} E_{i \rightarrow j},\ \textrm{where } e_i \in M_{stm}^{rem}
\end{gather*}

\item Using the engrams $M_{ltm}^{init}$ as a starting point, traverse the $M_{ltm}$ graph using the depth-first search (DFS) algorithm with a search depth of $N_{depth}$. The exploration direction should be based on the edge weight, toward the highest edge weight. Gather all the unique engrams that were encountered during the search, including $M_{ltm}^{init}$, and refer to them as $M_{ltm}^{found}$.

\begin{gather*}
    M_{ltm}^0 = M_{ltm}^{init} \\
    M_{ltm}^k = \argmax_{e_j \in M_{ltm}} E_{i \rightarrow j}, \\
    \textrm{ where } e_i \in M_{ltm}^{k-1},\ e_j \notin M_{ltm}^{found,k-1}\\
    M_{ltm}^{found,k} = \bigcup\limits_{l=0}^{k} M_{ltm}^l \\
    M_{ltm}^{found} = M_{ltm}^{found,N_{depth}}
\end{gather*}

\item Calculate correlation weight $C_{ltm}$ from $M_{wm}$ for $M_{ltm}^{found}$ and select top $N_{ltm}^{rem}$ number of engrams like STM. Denote the engrams as $M_{ltm}^{rem}$.

\item Use $M_{wm}, M_{stm}^{rem}, M_{ltm}^{rem}$ as activated memory.

\begin{equation*}
    \begin{aligned}
        M^{rem} &= M_{stm}^{rem} \cup M_{ltm}^{rem} \\
        M^{act} &= M_{wm} \cup M^{rem}
    \end{aligned}
\end{equation*}

\end{enumerate}

Cue-based activation is accomplished through a mechanism whereby only engrams with the highest correlation weight with working memory are finally activated.
This allows for effective memory searching without the requirement to access the entire long-term memory.
Rather, Memoria iteratively explores new engrams based on the engrams already discovered and their respective connection weights.

\subsection{Exploit}
\label{subsec:exploit}
In this phase, all the retrieved engrams are exploited to aid in task-solving and the contribution weight $w_{i}$ for each engram $e_i$ is evaluated. 
In our experiments, we considered the attention weight of each engram as the contribution, as engrams are referenced via the cross-attention mechanism.

\begin{figure*}[t!]
    \centerline{\includegraphics[width=\textwidth]{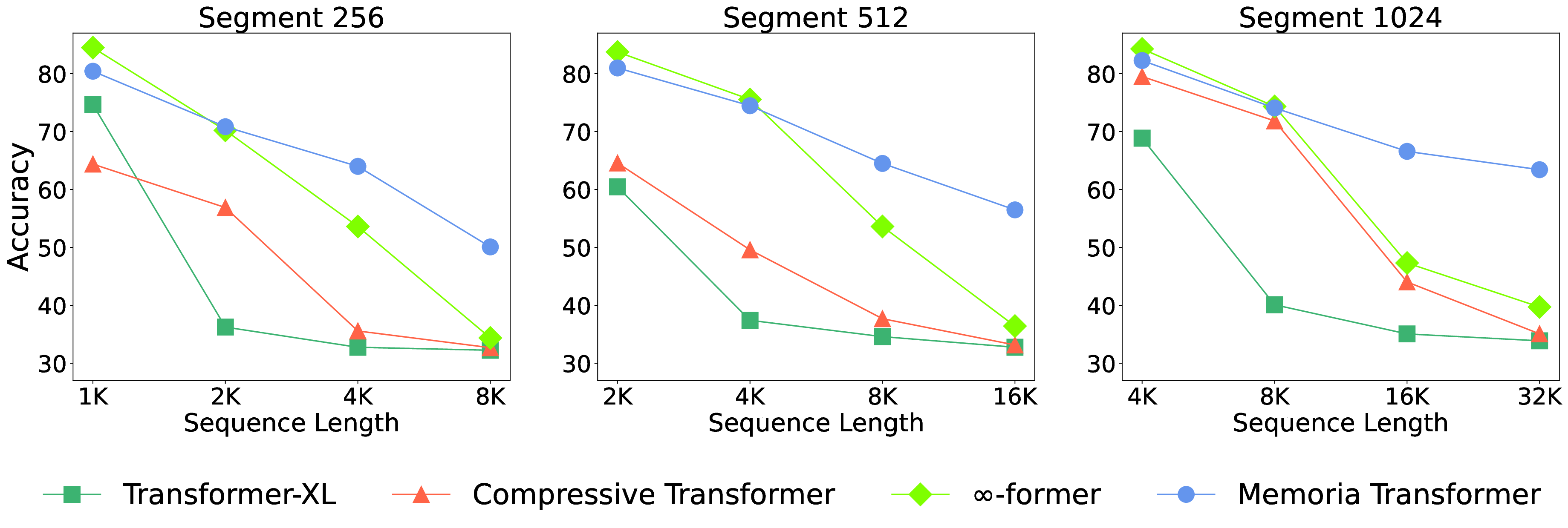}}
    \caption{Results of sorting task.
    Memoria Transformer exhibits greater robustness compared to other baselines as the input sequence length increases.
    This task requires the retention of information about the occurrence of initial tokens until the end.
    While the other methods all show significant performance decline, Memoria Transformer successfully handles the issue of fateful forgetting, setting it apart from other competing techniques.
    The comprehensive raw scores are specified in \cref{table:sorting-full-result}.
    }
    \label{fig:sorting-result}
\end{figure*}

\subsection{Memorize \& Forget}
\label{subsec:memorize-and-forget}
Along with achieving selective preservation, Memoria also strengthens the connections among associated engrams in this step.
\cref{fig:memorize-and-forget} shows the overall procedure of this stage.

\begin{enumerate}[itemsep=1ex,topsep=0ex,partopsep=0ex,parsep=0ex]
    \item Increase $Count_{i,j}$ by one for all engrams in $M^{act}$, which is the number of times $e_i$ and $e_j$ retrieved together. 
    \begin{gather*}
        \mathcal{N} = \{1, 2, \dots, |M^{act}|\} \\
        Count_{i,j} := Count_{i,j} + 1, \forall i, j \in \mathcal{N}
    \end{gather*}
    
    \item Increase lifespan of retrieved engrams by the increment ${Inc}_i$ for the engram $e_i$. ${Inc}_i$ is calculated as follows where $\alpha$ is hyperparameter meaning lifespan extend scale. If $\alpha$ is 1.0, each engram $e \in M^{rem}$ gets lifespan 1.0 on average.

    \begin{equation*}
        \begin{aligned}
            {Inc}_i = \frac{w_i}{\sum_{k=1}^{|M^{rem}|} w_k} \times |M^{rem}| \times \alpha
        \end{aligned}
    \end{equation*}
    
    \item Decrease lifespan of all engrams by 1.0.
    \item Remove engrams having a lifespan of 0 or less.
    \item Move $e_{wm}$ into STM. Reset WM.
    \item Move oldest engrams from STM by the number exceeding capacity into LTM.
\end{enumerate}

The differentiation in the lifespan occurs at two levels. First, engrams left unretrieved fail to attain lifespan and are prone to elimination. Retrieved engrams receive varying lifespans depending on their contribution, causing selective preservation at two levels. In addition, the connections among retrieved engrams are strengthened, facilitating co-retrieval of more closely associated engrams in memory searching.

%% file: pages/Memoria-Transformers.tex
Memoria functions independently as a module focused on managing engrams rather than directly participating in problem-solving process. Therefore, to effectively solve tasks using Memoria, it is powerful to merge it with neural network models. We integrated Memoria into two types of Transformers: a decoder-based model termed Memoria Transformer, and an encoder-based model referred to as Memoria BERT. Moreover, we employed a memory encoder module to generate engrams from the output of Transformer.

\begin{figure}[ht!]
    \centerline{\includegraphics[width=0.40\textwidth]{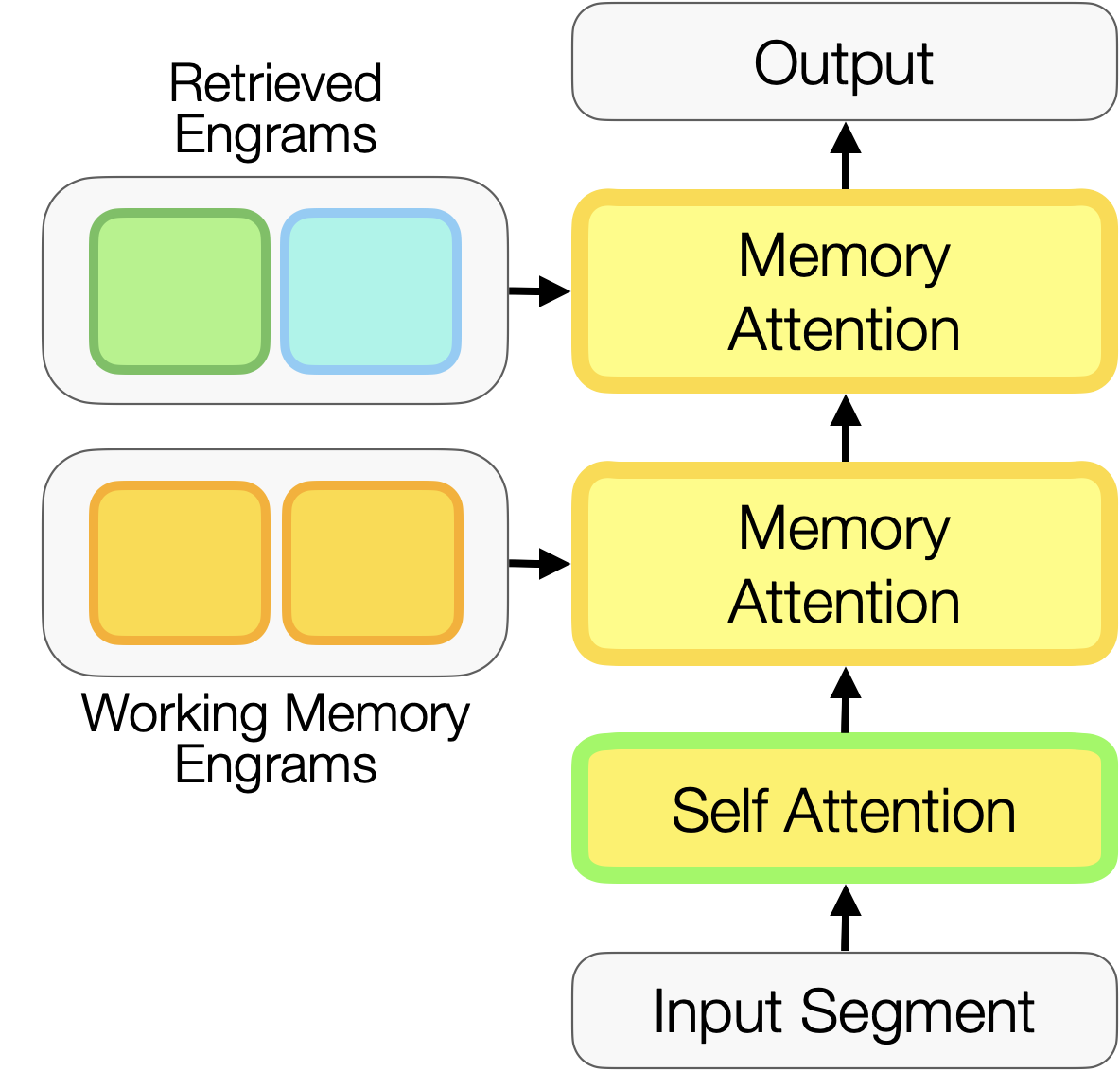}}
    \caption{A structural diagram of Memoria Transformers.}
    \label{fig:memoria-transformer-mini}
\end{figure}

Both Memoria Transformer and Memoria BERT reference engrams with the cross-attention mechanism. As illustrated in \cref{fig:memoria-transformer-mini}, the models first reference the working memory engrams and then the retrieved short-term/long-term memory engrams. The difference between the two models arises from how engrams are generated. Detailed architectures of each model are presented with description in \cref{sec:memoria-applied-transformers-structure}.

\textbf{Memoria Transformer}
We used the attention-based abstractor as memory encoder $f_e$ where queries are learnable parameters. As utilizing the information of the current time step leads to causal leakage, We used the last hidden state $h_{t-1}$ of the previous time step as $X_t$ in Memoria Transformer. $t$ represents the index of the segment in the whole sequence.
The three values of $Q$, $W_k$, and $W_v$ are trainable parameters. FFN is a feed-forward network as same in Transformer \citep{transformer}.
The number of working memory engrams $N_{wm}$ is determined by the number of queries $Q$, so the number of queries is a hyperparameter.

\begin{table*}[tb!]
\centering \small
\setlength{\tabcolsep}{3.5ex}
\caption{
Language modeling performance.
Perplexity (PPL) is provided for Wikitext-103 and PG-19 datasets, while bits-per-character (BPC) is shown for Enwik8.
In each case, the memory length matches the segment length, with Wikitext-103 and PG-19 using a length of 150, and Enwik8 using 512.
Memoria Transformer outperformed the other models on all datasets.
}
\begin{tabular}{lccc}
\toprule
Model & \multicolumn{1}{c}{Wikitext-103 (PPL)} & \multicolumn{1}{c}{PG-19 (PPL)} & \multicolumn{1}{c}{Enwik8 (BPC)} \\
\midrule
Transformer & 26.755 & 31.631 & 1.28 \\
Transformer-XL & 24.543 & 29.945 & 1.19 \\
Compressive Transformer & 24.794 & 29.603 & \textbf{1.16} \\
$\infty$-former & 24.685 & 29.154 & 1.21 \\
Memoria Transformer & \textbf{23.471} & \textbf{29.149} & \textbf{1.16} \\
\bottomrule
\end{tabular}
\label{table:language-modeling}
\end{table*}

\begin{figure*}[tb!]
    \centerline{\includegraphics[width=\textwidth]{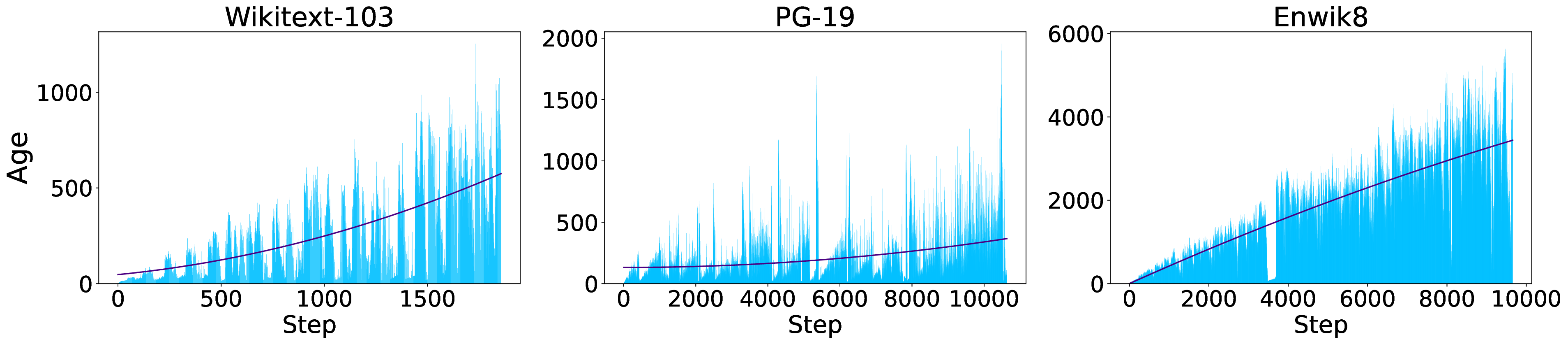}}
    \caption{The average age of retrieved engrams in long-term memory. The age of retrieved engrams gradually increased as steps passed by. This manifests that Memoria consistently retains and utilizes not only recent memories but also old ones.}
    \label{fig:retrieved-ltm-ages}
\end{figure*}

\begin{equation*}
    \begin{aligned}
        X_t &= h_{t-1} \\
        f_e(X_t) &= \textrm{Abstract}(X_t) \\
        &= \textrm{FFN}(\textrm{Attention}(Q, W_k X, W_v X)) \\
        &=  \textrm{FFN}(\textrm{Attention}(Q, W_k h_{t-1}, W_v h_{t-1})) \\
        &=  \textrm{FFN}(\textrm{softmax}(Q W_k h_{t-1}) W_v h_{t-1}) \\
        &= M_{wm}
    \end{aligned}
\end{equation*}

\textbf{Memoria BERT/RoBERTa} Memoria BERT also employs the same memory encoder as Memoria Transformer. Unlike decoder-based models, encoder-based models always have access to the complete input of the current time step without causing causal leakage.
Thus, memory encoder $f_e$ utilizes the hidden states $h_t^l$ as $X_t$, where $l$ signifies the memory layer index. New engrams are obtained from the hidden state of $l^{th}$ BERT layer through the abstractor. Subsequently, working memory engrams and retrieved engrams are employed in the following layers using cross-attention.

%% file: pages/Experiments.tex
We applied Memoria to Transformer and evaluated its ability to capture long-term dependencies in various tasks.
The first task is sorting. \citet{infformer} evaluated the model's ability to remember long-term information about the occurrence of numbers by generating a sorted sequence of numbers based on their predefined frequency of occurrence. Secondly, we performed language modeling for token-level on WikiText-103 (Raw) \citep{wikitext103} and PG-19 \citep{compformer}, and character-level on enwik8 \citep{enwik8}.
Similar to \citet{infformer}, only the first 2,000 books of the training dataset were used for PG-19. We compared Memoria with the other competitors of Transformer \citep{transformer}, Transformer-XL \citep{transfoxl}, Compressive Transformer \citep{compformer}, and $\infty$-former \citep{infformer}.
Lastly, we conducted the classification task on the long document classification dataset, Hyperpartisan \citep{hyperpartisan}. \cref{sec:additional-experiment} provides additional experiments and specifies the hyperparameters.

\subsection{Sorting}
\label{subsec:sorting-exp}
\input{pages/Experiments-Sorting}

\subsection{Language Modeling}
\label{subsec:language-modeling-exp}
\input{pages/Experiments-Language-Modeling}

\subsection{Classification}
\label{subsec:classification-exp}
\input{pages/Experiments-Classification}

%% file: pages/Experiments-Sorting.tex
The sorting task involves taking a sequence of symbols and outputting the symbols in descending order of frequency of occurrence \citep{infformer}. Decoder models including Memoria Transformer were utilized for this task.
We experimented with sequences of various lengths, ranging from 1K to 32K\footnote{We used the script of $\infty$-former at \url{https://github.com/deep-spin/infinite-former/blob/main/sorting/generate_data.py} to generate dataset.}, with segment lengths of 256, 512, and 1024, using the 20 unique tokens in the vocabulary.
In this task, it is essential to maintain the initial information until the end in order to avoid fateful forgetting, as the occurrence frequencies of a token vary from the beginning to the end.

\begin{table*}[t]
\centering \small
\setlength{\tabcolsep}{3.5ex}
\caption{Text classification performance on Hyperpartisan.
The evaluation metrics are average macro F1-score and accuracy, calculated through five independent runs. We report validation and test set results because of data distribution discrepancies.
}
\begin{tabular}{lcccc}
\toprule
\multirow{2}[2]{*}{Model [Sequence Length]} & \multicolumn{2}{c}{Validation} & \multicolumn{2}{c}{Test} \\
\cmidrule(lr){2-3}
\cmidrule(lr){4-5}
& \multicolumn{1}{c}{F1$_{\pm\text{STD}}$} & \multicolumn{1}{c}{Acc$_{\pm\text{STD}}$} & \multicolumn{1}{c}{F1$_{\pm\text{STD}}$} & \multicolumn{1}{c}{Acc$_{\pm\text{STD}}$} \\
\midrule
BERT [512] & $76.61_{\pm0.04}$ & $78.75_{\pm0.03}$ & $91.67_{\pm0.01}$ & $93.05_{\pm0.01}$ \\
RoBERTa [512] & $82.96_{\pm0.02}$ & $84.06_{\pm0.02}$ & $95.24_{\pm0.02}$ & $95.38_{\pm0.02}$ \\
Bigbird [4096] & $81.22_{\pm0.02}$ & $82.81_{\pm0.02}$ & $93.24_{\pm0.01}$ & $93.54_{\pm0.01}$ \\
Longformer [4096] & $78.33_{\pm0.03}$ & $79.69_{\pm0.03}$ & $94.56_{\pm0.01}$ & $94.77_{\pm0.01}$ \\
Memoria BERT [512] & $78.24_{\pm0.04}$ & $80.00_{\pm0.04}$ & $94.59_{\pm0.02}$ & $94.77_{\pm0.02}$ \\
Memoria RoBERTa [512] & $\textbf{86.39}_{\pm0.01}$ & $\textbf{87.19}_{\pm0.01}$ & $\textbf{96.51}_{\pm0.02}$ & $\textbf{96.62}_{\pm0.02}$ \\
\bottomrule
\end{tabular}
\label{table:hyperpartisan}
\end{table*}

\cref{fig:sorting-result} indicates the performance across different segment lengths in the sorting task as the sequence length expands. The memory length was set to the same value as the segment length. Generally, with increasing sequence length, accuracy tended to decline due to the necessity of retaining longer contextual information. Notably, Memoria exhibited the least degradation in performance compared to the other three models as sequence length increased, showcasing its ability to uphold long-term memory for extended context. 
To understand the respective roles of each memory and Hebbian property, we conducted ablation studies in \cref{sec:ablation-study}. This analysis verifies the complementary functions of each memory module and the significance of Hebbian property.

%% file: pages/Experiments-Language-Modeling.tex
\begin{table}[h!]
\centering \small
\setlength{\tabcolsep}{3.5ex}
\caption{Perplexity with a smaller segment length of 50.
Even in the shorter context and memory lengths, Memoria still maintained its superiority over the other methodologies.
}
\label{table:language-modeling-more-segment}
\begin{tabular}{lc}
\toprule
Model [Memory Length] & \multicolumn{1}{c}{Wikitext-103} \\
\midrule
Transformer & 39.287 \\
Transformer-XL [50] & 31.459 \\
Compressive Transformer [50] & 31.644 \\
$\infty$-former [50] & 31.790 \\
Memoria Transformer [48] & \textbf{30.007} \\
\bottomrule
\end{tabular}
\vspace{-0.1in}
\end{table}

In language modeling as well, Memoria Transformer was applied.
Since publicly available pre-trained models were trained with a varying number of parameters on different datasets, 
models were trained from scratch in our experiments.
Specifically, we employed GPT-2 architecture with 12 layers and 768 dimensions.
The results of additional experiments with pre-trained language models are detailed in \cref{subsec:language-modeling-detail}.
We set the segment length as 150 for token-level experiments and 512 for character-level experiments following the \citet{rmt}. 
The pre-trained GPT-2 tokenizer was used for all token-level experiments.

\cref{table:language-modeling} shows the results.
Compared to Transformer, all the alternative models exhibited enhanced performance.
Memoria Transformer achieved the best performance on all three datasets.
Such outcomes underscore the efficacy of Memoria in practical tasks given that language modeling involves complexities beyond merely capturing long-term context.

\cref{table:language-modeling-more-segment} presents the performance of each model when the segment length is reduced to 50, to observe the dynamics as the number of segments increases.
A comparison with \cref{table:language-modeling} highlights a more pronounced difference in performance between Transformer and memory models.
Even in situations demanding deeper consideration of long-term dependencies, Memoria steadily shows superior performance.

We validated whether Memoria effectively utilizes long-term memory. \cref{fig:retrieved-ltm-ages} shows the average age of the retrieved engrams in long-term memory at each step on the test dataset. The age represents the number of steps that have passed since the engram was created.
A flat line on the graph would suggest dependence only on recent engrams, rendering it ineffective as long-term memory. Conversely, continuous reference to past information causes engrams to age gradually, reflected in the graph's upward trend.
This trend signifies Memoria's consistent retrieval and utilization of important past information even after numerous time steps. 
For further clarity, several snapshots of internal connections are provided in \cref{sec:visualization-of-memoria} to aid comprehension.

%% file: pages/Experiments-Classification.tex
Hyperpartisan is a widely used dataset for the long document classification task. To validate the effectiveness of Memoria within encoder-based architectures, we integrated Memoria into BERT and RoBERTa and compared their performance with the other models.
Due to the high cost of pre-training models with different structures, it was inevitable to use pre-trained models for the classification task.
The size of all the models was 12-layer base-sized.
Memoria-augmented models referred to 192 engrams in addition to 512 context.

\cref{table:hyperpartisan} presents the classification performance of the models.
It is apparent that Memoria-augmented models show striking performance gains compared to plain BERT and RoBERTa, although it is challenging to compare the models altogether because of the disparity in pre-training.
Memoria RoBERTa achieved the highest score in all cases.
Conducting a one-tailed t-test, Memoria RoBERTa verified its statistically significantly higher performance than Longformer and Bigbird, with p-values of 0.045 and 0.005, respectively.

%% file: pages/Psychological-Memory-Effects.tex
\begin{figure}[ht!]
    \centerline{\includegraphics[width=0.445\textwidth]{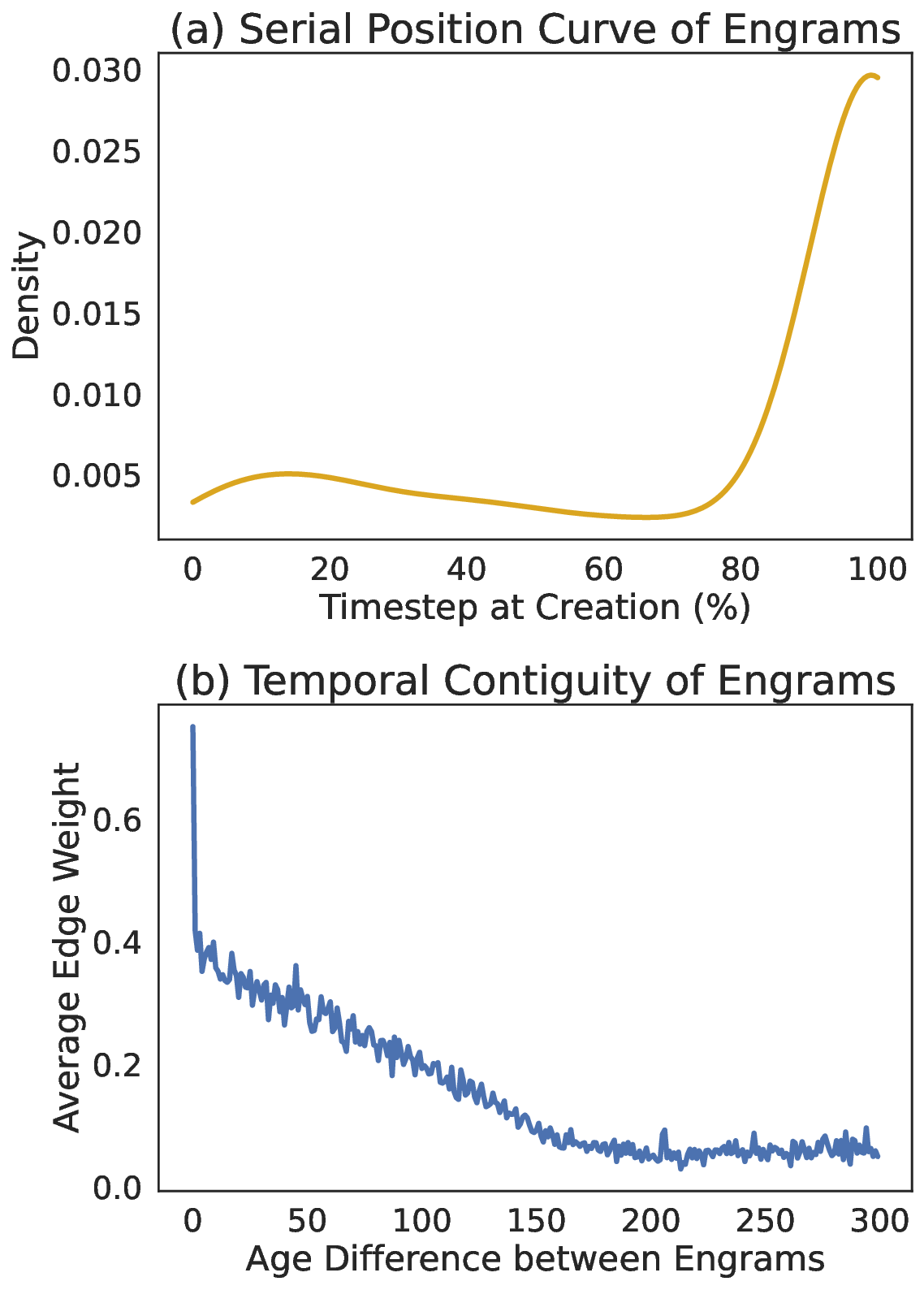}}
    \caption{Psychological memory effects of Memoria engrams.
    Figure (a) depicts a kernel density estimate plot illustrating the distribution of creation times for the remaining engrams in long-term memory after passing all time steps. This plot displays the patterns of the primacy effect and recency effect, where initial and recent information is more retained than intermediate information.
    Figure (b) illustrates the average connection weight according to the difference in age (time of creation) of the remaining engrams. This means that when the age difference is small, the edge weight is high, indicating the temporal contiguity effect. \cref{sec:psychological-memory-effects-details} provides further detailed explanations.
    } 
    \label{fig:psychological-effects}
\end{figure}

Memoria is designed based on the various memory models of humans. Particularly, the long-term memory of Memoria is deliberately engineered to maintain valuable old information, akin to humans. Due to the complexity of biological interpretations of human memory mechanisms, direct comparisons between Memoria and human memory are challenging. Nevertheless, research on the characteristics of human memory in psychology is extensive. Therefore, by identifying the known effects of human memory in Memoria, we confirm its resemblance to the human memory system.

\cref{fig:psychological-effects} exhibits patterns analogous to the primacy effect, recency effect, and temporal contiguity effect of engrams in Memoria. We utilized Memoria Transformer to conduct inference on the entire test set of Wikitext-103, subsequently analyzing the age and internal weights of remaining engrams.
The upper graph illustrates the density according to the timing of engram creation.
Conventional memory-based models are inclined to keep recent information, inducing an upward trend.
In contrast, Memoria preserves early and latest information more than intermediate information. This behavior demonstrates both the primacy and recency effect.

The lower figure shows the relationship between the age difference among the remaining engrams and the edge weight. Age represents the number of time steps that have passed since creation, and the age difference corresponds to the difference in creation time. We observed that engrams created closer in time exhibit higher internal connection weights, illustrating a pattern akin to the temporal contiguity effect.

%% file: pages/Conclusion.tex
We propose Memoria as a general memory module for neural networks, aiming to tackle the fundamental issue of fateful forgetting in long-term memory, along with the derived problems of long-term importance, selective preservation, cue-based activation, and memory searching.
The solutions to these problems draw inspiration from the human memory system, actively incorporating various psychological and neuroscientific theories related to memory, including the Multi-Store Model \citep{multistore-model}, SAM \citep{sam-origin}, and Hebbian theory \citep{Hebb1949}.
This approach allows Memoria to reflect various human memory effects, including the recency effect, primacy effect, and temporal contiguity effect. We demonstrate these effects through diverse analyses.
We validated Memoria's strong performance compared to the other methodologies in the tasks of sorting, language modeling, and classification.

We endeavored to empower Memoria with the powerful characteristics of human memory. However, discrepancies still exist in many aspects.
The levels of processing theory \citep{levels-of-processing} emphasizes a more continuous structure of memory based on the depth of processing rather than the discrete categories of the Multi-Store model.
Also, the interference theory \citep{interference} underscores the substantial impact of interference effects between established memories and incoming information as a predominant forgetting mechanism in long-term memory.

Our future research will incorporate these mechanisms into Memoria to align it more closely with the operational principles of human memory, augmenting its capabilities accordingly. 
We anticipate this integration will unleash the potential of Memoria in diverse sequential processing and agent-based tasks, particularly in fields such as conversational chatbot and reinforcement learning simulation, ultimately paving the way to realize human-level intelligence.

%% file: pages/Impact-Statement.tex
This paper seeks to create an external memory module designed for artificial neural networks to enhance the processing capabilities for general long-sequence data. The development of Memoria draws upon theories concerning human memory, which may have various societal implications. Notably, Memoria maintains information generated during inference in long-term memory. Consequently, prolonged use of specific user data for model inference might raise potential privacy concerns due to the accumulation of information in Memoria. Users of Memoria should carefully review regulations regarding the handling of personal data.

%% file: pages/appendix/Hebbian-attributes-of-memoria.tex
\citet{math-hebb} suggested six attributes of a useful plasticity model for Hebbian learning as follows. Memoria meets these attributes.

\paragraph{Locality}

The learning rule for the synapse $E_{i \rightarrow j}$ connecting neuron $j$ to neuron $i$ should depend only on the activity of $j$ and $i$ and not on the state of other neurons $k \neq i,j$.

\begin{equation*}
    \begin{aligned}
        E_{i \rightarrow j} &= \frac{Count_{i,j}}{Count_{i,i}} \\
    \end{aligned}
\end{equation*}

By definition, Memoria meets locality because it depends on only the count of $i,j$.

\paragraph{Cooperativity}

Hebb’s formulation ‘takes part in firing it’ implies that an increase in weight requires both the presynaptic and the postsynaptic neuron to be active.

\begin{equation*}
    \begin{aligned}
        E_{i \rightarrow j} \propto Count_{i,j} \\
    \end{aligned}
\end{equation*}

Since $E_{i \rightarrow j}$ is proportional to $Count_{i,j}$ and $Count_{i,i}$ never decreases, it only increases when $e_i$ and $e_j$ fire (retrieved) together.

\paragraph{Synaptic depression}

A mechanism for decreasing weights is a necessary requirement for any useful learning rule. There are three engrams $e_i, e_j, e_k$. $E_{i \rightarrow j}$ decreased when $e_i$ and $e_k$ fire together while $e_j$ does not. The superscript $pre$ means the value before firing of $e_i$ and $e_j$ and $post$ means the value after firing.

\begin{equation*}
    \begin{aligned}
        E_{i \rightarrow j}^{pre} &= \frac{Count_{i,j}^{pre}}{Count_{i,i}^{pre}} \\
        Count_{i,k}^{post} &= Count_{i,k}^{pre} + 1 \\
        Count_{i,i}^{post} &= Count_{i,i}^{pre} + 1 \\
        E_{i \rightarrow j}^{post} &= \frac{Count_{i,j}^{post}}{Count_{i,i}^{post}} \\
        &= \frac{Count_{i,j}^{pre}}{Count_{i,i}^{pre} + 1} \\
        &< E_{i \rightarrow j}^{pre}
    \end{aligned}
\end{equation*}

\paragraph{Boundedness}

In realistic rules, weights should remain bounded in a specific range. $E_{i \rightarrow j}$ must be between 0 and 1 because it is probability.

\begin{gather*}
    E_{i \rightarrow j} = P(e_j \in M^{rem} \mid e_i \in M^{rem}) \\
    0 \le P(e_j \in M^{rem} \mid e_i \in M^{rem}) \le 1
\end{gather*}

\paragraph{Competition}

The growth of some weights comes at the cost of a decrease in others. The increase of $E_{i \rightarrow j}$ requires the increase of $Count_{i,j}$ and $Count_{i,i}$. The increase of $Count_{i,i}$ reduces all the weight $E_{i \rightarrow k}$, for $ k \ne j$.

\paragraph{Long-term stability}

In adaptive systems, it is important to ensure that previously acquired knowledge is not forgotten. In Memoria, $E_{i \rightarrow j}$ is always the result of learning from all past examples because $Count$ is cumulative.

%% file: pages/appendix/Details-On-Psychological-Memory-Effects.tex
\subsection{Primacy and Recency Effects}

\begin{figure}[h]
    \centerline{\includegraphics[width=1.0\textwidth]{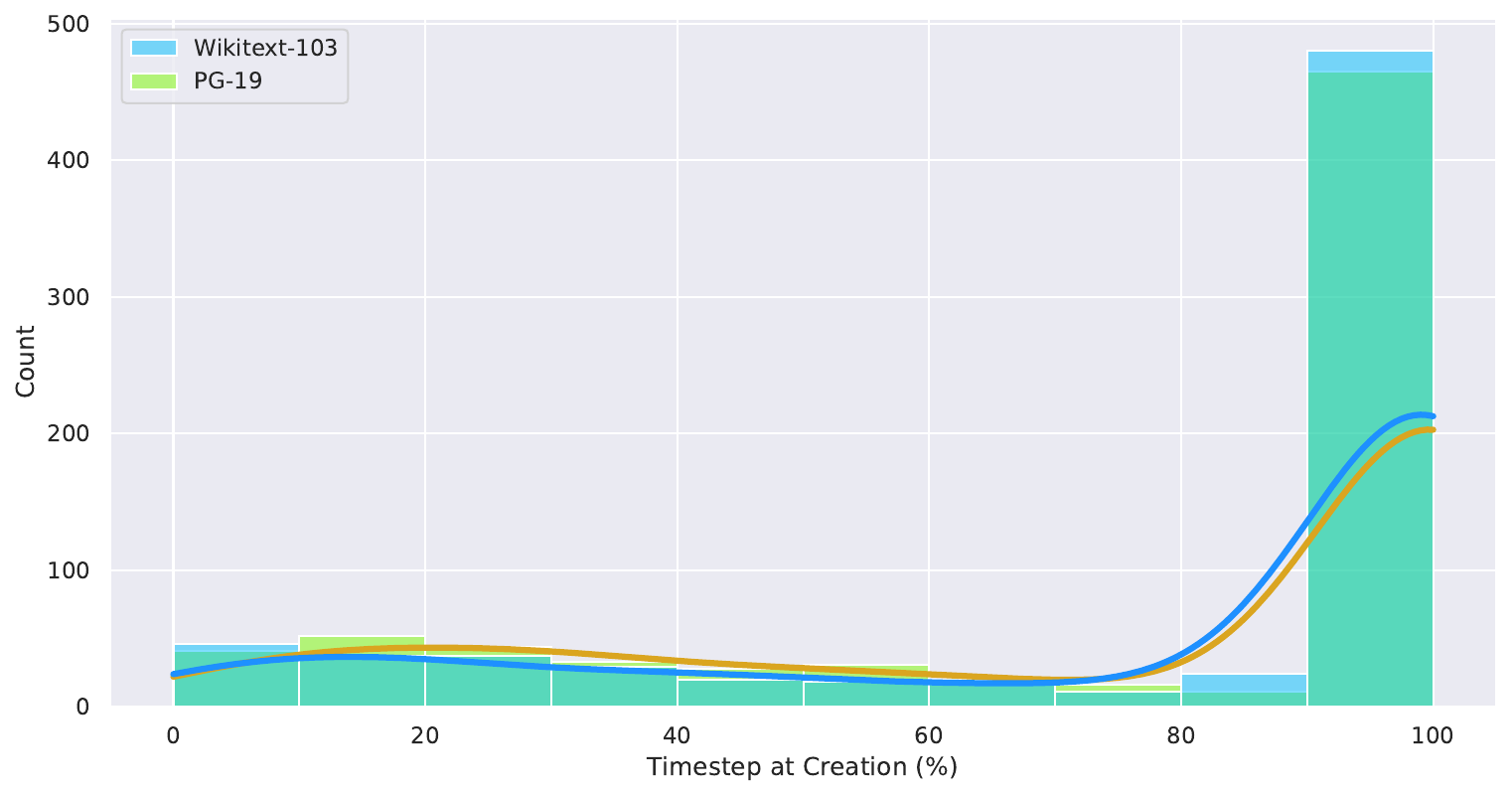}}
    \caption{Creation time of the remaining engrams in long-term memory after passing all time steps. The x-axis shows the time of creation as a percentage of the total time steps, and the y-axis is the count of engrams generated at that time. The trend line is the kernel density estimate. The graph illustrates the presence of both the primacy and recency effects, phenomena commonly observed in human memory, within Memoria as well.}
    \label{fig:engram-creation}
\end{figure}

Human memory, when processing sequentially arranged information, exhibits certain characteristics. Two of the most significant are the primacy effect \citep{serial-position, serial-position2} and the recency effect \citep{recency-effect, recency-effect2}.
The primacy effect is a cognitive bias that refers to the tendency of people to better remember and give greater importance to the first pieces of information, compared to information presented later. This phenomenon is particularly evident in situations where individuals are exposed to a series of items or stimuli, such as a list of words, a sequence of events, or a set of arguments.
The recency effect is a cognitive bias that refers to the tendency of individuals to better remember and give more weight to the most recent items or information in a series. In other words, when people are asked to recall a list of items, they are more likely to remember the items that appeared last in the list.
These effects can influence various aspects of memory and decision-making.
Both the recency and primacy effects are thought to be related to how information is processed and stored in memory.
The recency effect is believed to be influenced by short-term memory, where recently presented information is still readily available, while the primacy effect is associated with the transfer of information into long-term memory.

To verify whether these characteristics inherent in humans are also evident in Memoria, we analyzed engrams that persisted in long-term memory after passing all time steps, examining the point in time step when they were created.
\cref{fig:engram-creation} illustrates the number of remaining engrams at each creation time step.
As anticipated, the analysis results demonstrate the presence of both primacy and recency effects in Memoria. In the initial section, there is a clear indication of the primacy effect, as it exhibits a higher count compared to the middle portion. In the later part, there is a substantial difference, with a notably higher count compared to the middle, indicating the pronounced recency effect.
In Memoria, early created memories are likely to strengthen through subsequent retrieval. This resembles the primacy effect in humans, where early information tends to be well-maintained due to frequent chances for rehearsal \citep{rehearsal-process-in-free-recall}.
More recent memories have a high probability of being preserved due to their remaining lifespan, causing the recency effect.
Additionally, the observed distribution of engrams across the overall time steps in Memoria indicates successful mitigation of the initial problem of fateful forgetting, which was a key objective of our study. 

\subsection{Temporal Contiguity Effect}

\begin{figure}[h!]
    \centerline{\includegraphics[width=0.75\textwidth]{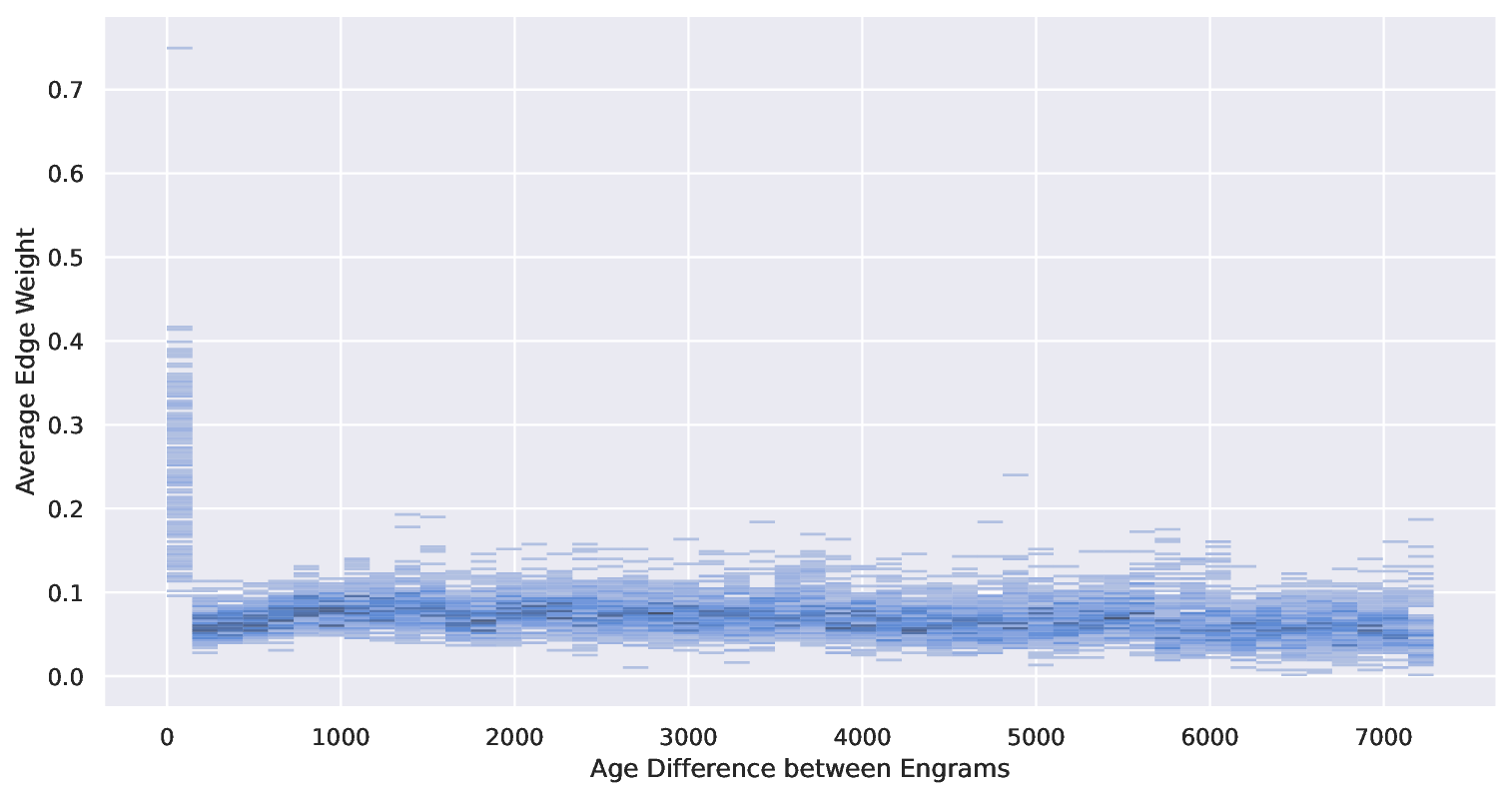}}
    \caption{A bivariate histogram showing the average weight between all engrams according to the age difference.}
    \label{fig:weights-per-age-diff}
\end{figure}

\begin{figure}[h!]
    \centerline{\includegraphics[width=0.75\textwidth]{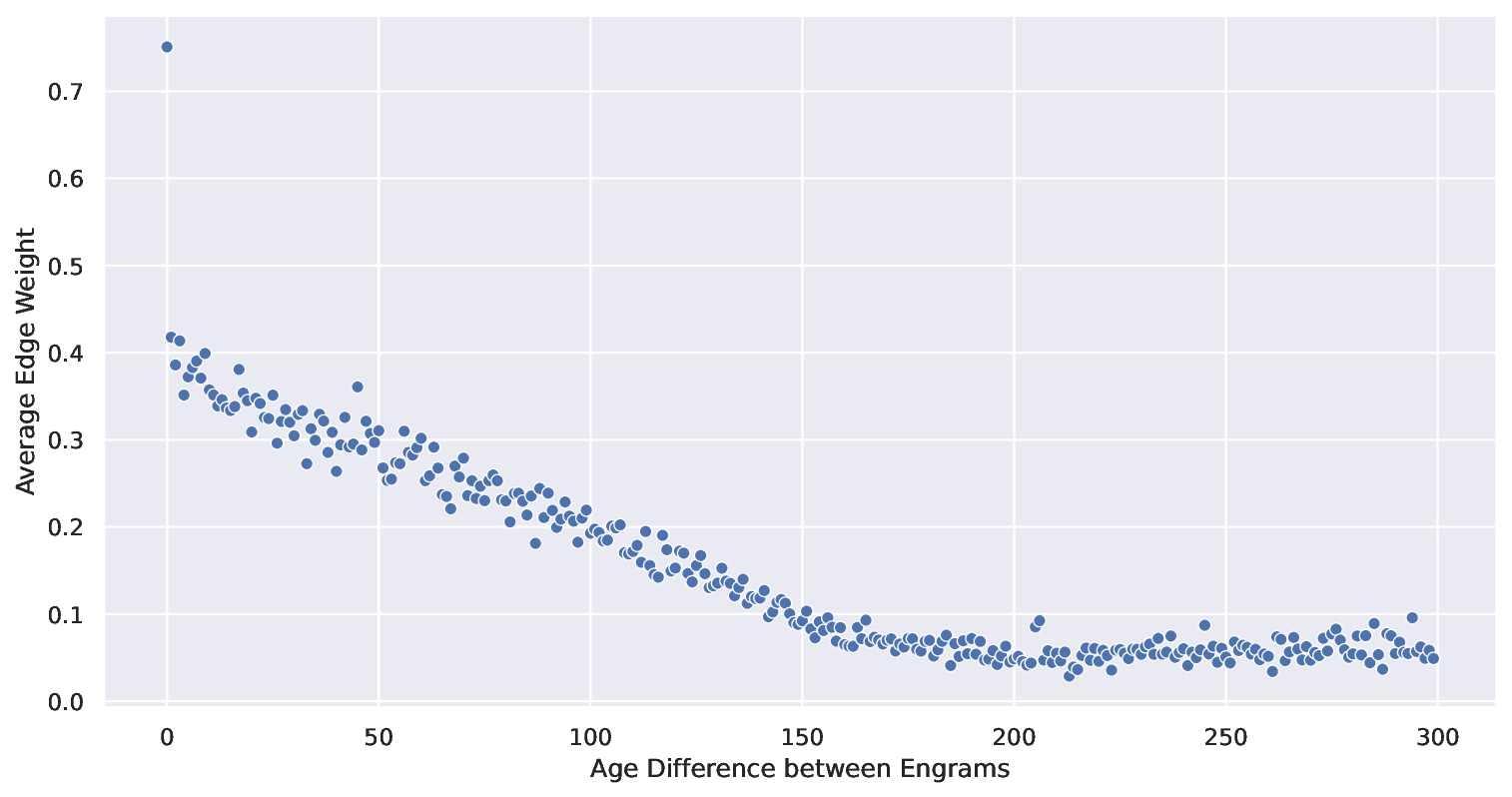}}
    \caption{Scatterplot of the data in \cref{fig:weights-per-age-diff} for age differences of 300 or less.}
    \label{fig:weights-per-age-diff-scatter}
\end{figure}

The temporal contiguity effect \citep{temporal-contiguity-effect, temporal-contiguity-effect2} is a cognitive phenomenon that enhances memory recall and comprehension when information elements are presented in close temporal proximity. In other words, humans tend to remember and understand information better when it is presented in a temporally aligned manner, as opposed to when there are temporal gaps between different elements. This effect facilitates the understanding of relationships between events occurring within a short time span.

To investigate the occurrence of the temporal contiguity effect in Memoria, we conducted an analysis of the connectivity among the remaining engrams in the WikiText-103 test dataset, after passing all time steps. We present a depiction of the changes in the average edge weight, which is based on the age difference (or the temporal gap at the time of creation) between two engrams, in \cref{fig:weights-per-age-diff}. The results distinctly show that a minimal age difference leads to a significant increase in the average weight. This suggests the existence of the temporal contiguity effect in Memoria, which strengthens the connections between engrams that are close in time. \cref{fig:weights-per-age-diff-scatter} provides a detailed representation of the point at which this effect begins to fade, indicating its presence up to approximately 150 age differences. Theoretically, engrams of the same age have a higher likelihood of firing together when they exist in working memory and short-term memory, resulting in an increase in the association edge between them.

\subsection{Retrieval Practice Effect}

\begin{table}[htb!]
\centering \small
\setlength{\tabcolsep}{3.5ex}
\caption{Autocorrelation coefficients of short-term memory and long-term memory engrams.}
    \begin{tabular}{lcc}
        \toprule
        Lag & Short-term Memory ACF & Long-term Memory ACF \\
        \midrule
        1 & 0.900 & 0.575 \\
        2 & 0.893 & 0.529 \\
        3 & 0.889 & 0.501 \\
        4 & 0.888 & 0.475 \\
        5 & 0.888 & 0.461 \\
        6 & 0.890 & 0.442 \\
        7 & 0.893 & 0.426 \\
        8 & - & 0.413 \\
        9 & - & 0.395 \\
        10 & - & 0.381 \\
        11 & - & 0.370 \\
        12 & - & 0.356 \\
        13 & - & 0.344 \\
        14 & - & 0.333 \\
        15 & - & 0.321 \\
        \bottomrule
    \end{tabular}
\vspace{0.1in}
\label{table:acf}
\end{table}

We conducted autocorrelation analysis to understand the association between the retrieval of each engram and subsequent retrieval events using Wikitext-103 dataset.
\cref{table:acf} presents the autocorrelation coefficients for short-term and long-term memory. We encoded retrieval of an engram as 1 and non-retrieval as 0. Lag represents the time step difference for correlation calculation. For instance, the lag of one signifies the autocorrelation between retrieval of engram $e_i$ at time $t$ and retrieval at time $t+1$. 
We obtained individual correlation coefficients from each engram, then aggregated them by lag and computed the average.
For short-lived engrams, with a tendency to be always retrieved or always not, most of those engrams have variances of 0. We regarded the coefficient of these cases as one because its actual meaning is a very strong autocorrelation. In addition, for long-term memory, we calculated the weighted average of the correlation coefficients in proportion to the lifespan of each engram, as the total lifespan differs for each engram.

\begin{figure}[bh!]
    \centerline{\includegraphics[width=\textwidth]{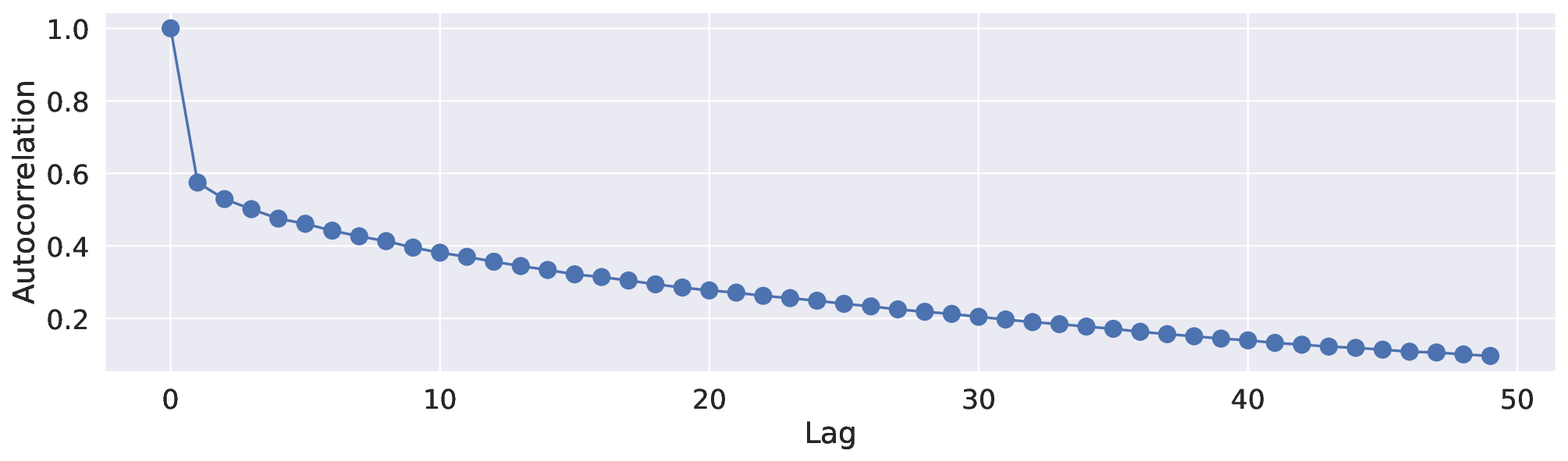}}
    \caption{Autocorrelation coefficient plot of long-term memory.}
    \label{fig:acf-ltm}
\end{figure}

First, looking at short-term memory, the capacity of short-term memory is 400, so each memory stays in short-term memory for 8 time steps. Therefore, the maximum observable lag is 7. Each engram in short-term memory has a significantly high autocorrelation. This implies that once an engram is retrieved, it is easy for it to be retrieved again, indicating that a specific memory is more frequently associated with others. Long-term memory also shows significant autocorrelation, displaying a strong correlation in close time intervals that gradually diminishes over extended periods. Theoretically, once an engram in long-term memory is retrieved, the association with more recent memories strengthens, making the old memory easier to reach through the pathway of those recent memories.
In human memory, a retrieval event itself makes the retrieved information more retrievable in the future \citep{retrieval-as-memory-modifier, retrieval-as-self-limiting-process, retrieval-practice}, a phenomenon known as the testing effect or retrieval practice effect.
The high autocorrelation of retrieval events in Memoria partially implies the manifestation of such phenomena.
\cref{fig:acf-ltm} illustrates the changes in autocorrelation based on the lag in long-term memory.

%% file: pages/appendix/Ablation-Study.tex
\subsection{Memory Type}

\begin{table}[h]
\centering \small
\setlength{\tabcolsep}{3.5ex}
\caption{Performance and performance gain of each memory module according to the length of the dataset.
Memoria consistently exhibits excellent performance, even as the sequence length increases. This is attributed to the harmonious interplay of each memory module. This observation indicates that while the effectiveness of the working memory diminishes with extended sequence lengths, both short-term and long-term memory modules display an enhanced performance.
}
    \begin{tabular}{lrrrrr}
        \toprule
        ~ & 4K & 8K & 16K & 32K & 48K \\
        \midrule
        Number of Segments & 4 & 8 & 16 & 32 & 47 \\
        \midrule
        \textit{Accuracy} & & & & & \\
        Transformer & 36.19 & 33.79 & 31.69 & 29.94 & 19.04 \\
        + Working Memory & 79.69 & 70.85 & 62.21 & 52.01 & 34.32 \\
        + Short-term Memory & 82.66 & 76.20 & 66.37 & 58.75 & 54.87 \\
        + Long-term Memory & 82.27 & 74.08 & 66.58 & 63.42 & 63.26 \\
        \midrule
        \textit{Performance Gain} & & & & & \\
        + Working Memory & +43.50 & +37.06 & +30.52 & +22.07 & +15.28 \\
        + Short-term Memory & +2.79 & +5.35 & +4.16 & +6.74 & +20.55 \\
        + Long-term Memory & -0.39 & -2.12 & +0.21 & +4.67 & +8.39 \\
        \bottomrule
    \end{tabular}
\vspace{0.1in}
\label{table:ablation}
\end{table}

We executed an ablation study to scrutinize the influence of each memory type in Memoria on performance. This study, conducted on a sorting task with a fixed segment length of 1024 for each dataset, enabled us to observe trends relative to data length. To extend our investigation to longer datasets not discussed in the main text, we performed supplementary experiments with a 48K dataset. Since the segment length remains constant at 1024, an expansion in the dataset length consequently increases the number of segments.

The analysis results indicate that each type of memory module contributes to overall performance to some extent. An noteworthy observation is that as the number of segments increases, the influence of each type of memory on performance changes. A closer look at the 4K dataset results, with a mere segment length of 4, reveals that the bulk of performance enhancement is primarily driven by the working memory.
However, as the dataset length expands to 8K, 16K, and beyond, the performance boost attributed to the working memory rapidly declines. In contrast, with the extension of sequence lengths, the impact of both short-term and long-term memory on performance progressively intensifies.
This trend mirrors human memory, which varies in retention duration depending on the situation.

This trend suggests that the model does not employ all types of memory uniformly, but rather selectively uses memory information based on the task or dataset characteristics. If the task can be sufficiently handled with a comprehension of short contexts, the model predominantly utilizes working memory. However, when confronted with longer contexts that pose a challenge for the working memory alone, the model appears to cultivate the capacity to harness short-term or long-term memory. Notably, when observing the transition from 32K to 48K, it becomes clear that the final performance difference between 32K and 48K is negligible when all memories are engaged, owing to the complementary roles of short-term and long-term memory compensating for the further aggravated performance deficiencies in the Transformer or working memory. These insights imply that future tasks and datasets should sufficiently demand dependency on long-term context to effectively validate the model’s long-term memory capabilities. Memoria consistently exhibits robust performance across datasets of diverse lengths, thanks to the complementary roles of the three types of memory.

\subsection{Hebbian Property}

\begin{table}[htb!]
\centering \small
\setlength{\tabcolsep}{3.5ex}
\caption{Performance and performance gain when modifying Hebbian characteristics.
This outcome underscores that weight modification, grounded in Hebbian theory within Memoria, yields markedly superior performance enhancements as opposed to arbitrary weight increments.
Furthermore, Memoria showcases its efficiency in preserving a substantial portion of performance achieved when conducting a full search of the entire long-term memory.
}
    \begin{tabular}{lrr}
        \toprule
        ~ & Accuracy & Performance Gain \\
        \midrule
        Memoria Transformer (Without LTM) & 54.87 & - \\
        \midrule
        Memoria Transformer & 63.26 & +8.39 \\
        Memoria Transformer (Random Wire) & 56.13 & +1.26 \\
        Memoria Transformer (Full LTM Search) & 65.84  & +10.97 \\
        \bottomrule
    \end{tabular}
\vspace{0.1in}
\label{table:ablation-hebbian}
\end{table}

We carried out further experiments to examine the efficiency and effectiveness of weight adjustments, guided by Hebbian theory.
This experiment was conducted using a 48K length dataset in a sorting task, for which the performance gain of long-term memory is established from a prior ablation study.
\cref{table:ablation-hebbian} presents the experimental results.

Firstly, the term `Random Wire' refers to a condition modified by Memoria to randomly select target engrams. Memoria strengthens connections between all working memory and retrieved engrams, denoted as $M^{act}$. Thus, it increases all $Count_{i,j}$ satisfying $e_i \in M^{act}$ and $e_j \in M^{act}$. In contrast, under the random wire condition, it increases $Count_{i,j}$ satisfying $e_i \in M^{act}$ and $e_j \in M_{ltm}$.
The total increase in count was controlled. The experimental findings indicate that substituting Hebbian process with random wiring results in a notable reduction in the performance gain of long-term memory, reducing from 8.39 to 1.26, which is approximately one-sixth of the original value. These results highlight the substantial effectiveness of weight adjustments following Hebbian process on performance.

The term `Full LTM Search' refers to selecting the highest engrams by simply calculating $C_{ltm}$ across all long-term memory without employing the internal weights in Memoria. In simpler terms, it treats $M_{ltm}$ as $M_{ltm}^{found}$. In this case, there was a performance enhancement of 2.58. These findings demonstrate that Memoria's long-term memory searching preserves a substantial portion of the performance while utilizing significantly fewer computing resources than the naive approach.

%% file: pages/appendix/Training-Details-And-Additional-Results.tex
For all experiments, the Adam optimizer and linear scheduler with warm-up were used, and the gradient clipping was set to a norm of 1.0. One or more NVIDIA A100 or A6000 GPUs were used for training.

\subsection{Sorting}
\label{subsec:sorting-detail}

For all sorting experiments, a batch size of 32, a warmup rate of 0.06, a learning rate of 2e-4, and an epoch of 5 were used for 80,000 train examples. A memory length was configured to match the segment length. The experiments were conducted on datasets with lengths ranging from 1000 to 32,000. Each example on the datasets was divided into segments of lengths 256, 512, and 1024. For each segment length, combinations of sequence lengths and segment lengths were constructed by varying the number of segments, which were set to 4, 8, 16, and 32. The model configuration used was 5 layers, 4 heads, embedding dimension of 512 Transformer. The compression rate is 4 and the ratio of normal memory and compressed memory is one-to-one for Compressive Transformer.

Memoria parameters used in the experiment were as follows: an initial lifespan of 5, a lifespan extension scale $\alpha$ of 8, and a long-term memory search depth $N_{depth}$ of 10 in all cases. Other parameters are adjusted proportionally to the segment length. the number of working memories $N_{wm}$ set to 1/8 of the segment length, the number of retrieved engrams in short-term memory $N_{stm}^{rem}$ set to 1/4 of the segment length, the number of retrieved engrams in long-term memory $N_{ltm}^{rem}$ set to 5/8 of the segment length, and a capacity of short-term memory set to half of the segment length. The sum of $N_{wm}$, $N_{stm}^{rem}$, and $N_{ltm}^{rem}$ is equal to the segment length.

\cref{table:sorting-full-result} shows the all scores of models in the sorting task. The metric is accuracy. For the convenience of comparison, we mark the number of segments instead of the total sequence length of each dataset. The sequence length can be obtained by multiplying the number of segments by segment length. Memoria Transformer proves its robustness for long-term dependency compared to the other models, especially as the number of segments increases.

\begin{table}[htb!]
\centering \small
\setlength{\tabcolsep}{3.5ex}
\caption{Accuracy in the sorting task. When the segments increase, Memoria outperforms other baselines.}
\begin{tabular}{lcccc}
\toprule
\multirow{2}[2]{*}{Model} & \multirow{2}[2]{*}{Segments} & \multicolumn{3}{c}{Segment Length} \\
\cmidrule(lr){3-5}
&& 256 & 512 & 1024 \\
\midrule
Transformer-XL & 4 & 74.66 & 60.46 & 68.86 \\
Compressive Transformer & 4 & 64.38 & 64.57 & 79.51 \\
$\infty$-former & 4 & \textbf{84.49} & \textbf{83.75} & \textbf{84.28} \\
Memoria Transformer & 4 & 80.42 & 80.99 & 82.27 \\
\midrule
Transformer-XL & 8 & 36.24 & 37.41 & 40.09 \\
Compressive Transformer & 8 & 56.88 & 49.58 & 71.84 \\
$\infty$-former & 8 & 70.21 & \textbf{75.55} & \textbf{74.34} \\
Memoria Transformer & 8 & \textbf{70.84} & 74.47 & 74.08 \\
\midrule
Transformer-XL & 16 & 32.75 & 34.59 & 35.06 \\
Compressive Transformer & 16 & 35.57 & 37.69 & 44.03 \\
$\infty$-former & 16 & 53.61 & 53.61 & 47.31 \\
Memoria Transformer & 16 & \textbf{63.99} & \textbf{64.50} & \textbf{66.58} \\
\midrule
Transformer-XL & 32 & 32.24 & 32.76 & 33.87 \\
Compressive Transformer & 32 & 32.68 & 33.15 & 35.07 \\
$\infty$-former & 32 & 34.36 & 36.41 & 39.71 \\
Memoria Transformer & 32 & \textbf{50.08} & \textbf{56.48} & \textbf{63.42} \\
\bottomrule
\end{tabular}
\vspace{0.1in}
\label{table:sorting-full-result}
\end{table}

\subsection{Language Modeling}
\label{subsec:language-modeling-detail}

For all language modeling experiments, a batch size of 8 and a warmup rate of 0.06 were used. The model configuration used the settings of GPT-2 small by default. The Wikitext-103 and PG-19 datasets were trained for 3 epochs, while the Enwik8 dataset was trained for 20 epochs. GPT-2 tokenizer was used for all datasets except Enwik8, which was trained at the character level using 204 characters. The default learning rate was 2e-4, but in cases where convergence was challenging, 1e-4 was used. However, for experiments fine-tuning pre-trained models, a learning rate of 5e-5 was used. In the experiments conducted on the Wikitext-103 dataset using Transformer-XL and on the PG-19 dataset using  $\infty$-former, as well as the experiment with reduced segment length to 50, both Memoria Transformer and Transformer-XL were trained with a learning rate of 1e-4. The memory length was set to be the same or similar to the segment length. The compression rate is 4 and the ratio of normal memory and compressed memory is one-to-one for Compressive Transformer.

Memoria parameters were set as follows: initial lifespan of 9, lifespan extend scale $\alpha$ of 8, and long-term memory search depth $N_{depth}$ of 10. Furthermore, to prevent potential interference with the learning process, we periodically reset all memory in Memoria every 500 steps during training (1500 steps for enwik8 dataset). This was done to avoid referencing memory generated at stages where learning was insufficient, as it could impede the training progress. For the Wikitext-103 and PG-19 datasets, the number of working memories $N_{wm}$, the number of retrieved engrams in short-term memory $N_{stm}^{rem}$, and the number of retrieved engrams in long-term memory $N_{ltm}^{rem}$ were all set to 50, and a capacity of short-term memory was set to 400. For the Enwik8 dataset, $N_{wm}$, $N_{stm}^{rem}$ and $N_{ltm}^{rem}$ were set to 170, and a capacity of short-term memory was set to 1360. When training on the Wikitext-103 dataset with a reduced segment length of 50, $N_{wm}$, $N_{stm}^{rem}$, and $N_{ltm}^{rem}$ were all set to 16, and the short-term memory capacity was set to 128.

\begin{table}[htb!]
\centering \small
\setlength{\tabcolsep}{3.5ex}
\caption{Finetuning performance on Wikitext-103.}
\begin{tabular}{lc}
\toprule
Model & Wikitext-103 \\
\midrule
GPT-2 & 20.498 \\
Memoria GPT-2 & \textbf{18.986} \\
\midrule
GPT-2 Large & 15.332 \\
Memoria GPT-2 Large & \textbf{13.227} \\
\midrule
GPT-2 XL & 15.254 \\
Memoria GPT-2 XL & \textbf{13.241} \\
\bottomrule
\end{tabular}
\vspace{0.1in}
\label{table:language-modeling-finetune}
\end{table}

To verify whether Memoria can consider long-term context even when finetuning a pre-trained model, we measured performance on Wikitext-103 dataset by finetuning Memoria GPT-2. The architecture of Memoria GPT-2 is the same as Memoria Transformer. The results are \cref{table:language-modeling-finetune}. Memoria GPT-2 showed significantly better performance than GPT-2. This result suggests that Memoria can be combined with various pre-trained models to increase long-term dependencies. Furthermore, as the use of pre-trained large language models (LLMs) has become prevalent, we conducted experiments to verify whether Memoria can be applied in conjunction with LLMs. We performed experiments using large and xl sized models, and successfully achieved performance improvements when applying Memoria to even larger pre-trained models. This demonstrates the potential for LLMs to benefit from considering longer contexts with the help of Memoria.

\begin{table}[!ht]
\centering \small
\caption{Performance on Wikitext-103 under various parameter variations.}
\begin{tabular}{lllllll|l}
\toprule
Initial lifespan & Lifespan extend scale $\alpha$ & LTM search depth & Reset Period & $N_{wm}$ & $N_{stm}^{rem}$ & $N_{ltm}^{rem}$ & Perplexity \\
\midrule
9 & 8 & 10 & 500 & 50 & 50 & 50 & 23.471 \\
\midrule
5 & ~ & ~ & ~ & ~ & ~ & ~ & 23.485 \\
\midrule
~ & 4 & ~ & ~ & ~ & ~ & ~ & 23.518 \\
\midrule
~ & ~ & 5 & ~ & ~ & ~ & ~ & 23.491 \\
\midrule
~ & ~ & ~ & 100 & ~ & ~ & ~ & 23.407 \\
\midrule
~ & ~ & ~ & ~ & 100 & 25 & 25 & \textbf{23.376} \\
\midrule
~ & ~ & ~ & ~ & 25 & 100 & 25 & 23.831 \\
\midrule
~ & ~ & ~ & ~ & 25 & 25 & 100 & 23.670 \\
\bottomrule
\end{tabular}
\label{table:language-modeling-param-variations}
\end{table}

We conducted additional experiments by modifying various hyperparameters to investigate the sensitivity of hyperparameters. \cref{table:language-modeling-param-variations} shows that there is generally no significant difference in performance compared to what was initially reported in the paper. There were even cases where the performance improved compared to the original scores.

\subsection{Classification}
\label{subsec:classification-details}

All hyperpartisan text classification experiments were conducted with a batch size of 16, a learning rate of 5e-5, and a warmup rate of 0.1. The models were trained for 20 epochs. For BERT, the experiment utilized the pre-trained bert-base-uncased model. As for Longformer, the base model was used in the experiment.

Memoria parameters used in the experiment were as follows: an initial lifespan of 12, a lifespan extension scale $\alpha$ of 8, a long-term memory search depth $N_{depth}$ of 10, the number of working memories $N_{wm}$ set to 64, the number of retrieved engrams in short-term memory $N_{stm}^{rem}$, and the number of retrieved engrams in long-term memory $N_{ltm}^{rem}$ both set to 64, a capacity of short-term memory of 128, and the memory layer index set to 9. This means that the output of the 10th layer is used as memory, and it is referenced in the remaining 2 layers of the model.

%% file: pages/appendix/Algorithm-And-Computational-Complexity.tex

\subsection{Theoretical Analysis}
\label{subsec:theoretical-analysis}

Each stage of Memoria is represented by an algorithm. These are the algorithms of decoder models in our experiments, so some details might be slightly different from the encoder model's formula. Additionally, each algorithm provides time complexity to help estimate how many resources are needed.

\begin{algorithm}
    \caption{Retrieve Stage}
\begin{algorithmic}
    \STATE {\bfseries Input:} short-term memory $STM$, long-term memory $LTM$, memory encoder $E$, co-retrieved conditional probabilities $P$, previous hidden states $h_p$, long-term memory search depth $N_{depth}$
    \STATE {\bfseries Output:} {working memory $WM$, retrieved engrams $retrieved$}
    \STATE {\bfseries Result:} {Encode $h_p$ into working memory. Find relevant engrams in the short-term/long-term memories.}
    
    \STATE $WM \gets E(h_p)$
    \STATE $W_{stm} \gets CalculateDistance$($STM$, $WM$)
    \COMMENT{distance from stm to wm}
    \STATE $stm_{rem} \gets FindShortestK$($W_{stm}$)
    \COMMENT{select nearest stms}
    \STATE $p \gets GetCondProb$($LTM$, $stm_{rem}$, $P$)
    \STATE $ltm_{1} \gets SelectMostProbableEngrams$($p$)
    \STATE $ltm_{found} \gets (ltm_{1},)$\;
    
    \FOR{$i \gets 1$ {\bfseries to} $N_{depth}$}
        \STATE $p \gets GetCondProb$($LTM$, $ltm_i$, $P$)
        \STATE$ltm_{i+1} \gets SelectMostProbableEngrams$($p$)
        \STATE $Append$($ltm_{found}$, $ltm_{i+1}$)
    \ENDFOR
    \STATE $W_{ltm} \gets CalculateDistance$($ltm_{found}$, $WM$)
    \STATE $ltm_{rem} \gets FindShortestK$($W_{ltm}$)
    \STATE $retrieved \gets Merge$($stm_{rem}$, $ltm_{rem}$)
\end{algorithmic}
\end{algorithm}

The complexity of the $CalculateDistance$ function is equal to the product of the number of engrams in each memory, as it involves the computation of all weights between them. The function is used twice, first in the STM with a time complexity of $O(N_{wm} \times C_{stm})$, where $N_{wm}$ is the number of engrams in working memory and $C_{stm}$ is the capacity of STM. Secondly, when applied to the found LTM, the complexity is $O(N_{wm} \times N_{ltm}^{found})$, where $N_{ltm}^{found} = N_{stm}^{rem} \times (N_{depth} + 1)$.
The part of the function that retrieves the conditional probability of retrieving the connected LTM engrams given retrieved STM engrams has a complexity of $O(N_{stm}^{rem} \times d)$, where $N_{stm}^{rem}$ is the number of retrieved engrams in STM and $d$ is the degree. The maximum value for degree $d$ is the total number of edges from the engram, resulting in a maximum complexity of $O(N_{stm}^{rem} \times N_{ltm})$.
Within the loop that executes $N_{depth}$ times, the complexity is $O(N_{stm}^{rem} \times N_{ltm} \times N_{depth})$. Generally, since the size of LTM is expected to be larger than $N_{wm}$, the overall time complexity of the retrieve stage is $O(N_{stm}^{rem} \times N_{ltm} \times N_{depth})$.

Here, $N_{stm}^{rem}$ and $N_{depth}$ are hyperparameters that can be set directly, but the total number of long-term memory units, $N_{ltm}$, is a dynamically changing value during execution. While it is not possible to precisely determine the size of LTM, the maximum size of LTM over time can demonstrate convergence through lifespan, given a sufficient duration. The increase in lifespan for all engrams during a single execution of the entire memory operations is $\alpha * (N_{stm}^{rem} + N_{ltm}^{rem})$ when alpha represents the lifespan extend scale parameter. Additionally, the decrease in lifespan is the number of all engrams of $N_{ltm} + N_{stm} + N_{wm}$.
In a scenario where $N_{ltm}$ is maximized, lifespan is evenly distributed across all engrams, preventing their removal. If the sum of lifespans for all engrams after the nth execution is denoted as $l$, then $N_{ltm}$ can be considered a constant multiple, $l \times c$. However, since the total number of engrams cannot exceed the total lifespan sum, $c$ takes on values between 0 and 1. When memory operations are executed $n$ times, and the total lifespan sum of all engrams is $l_n$, $l_n$ can be expressed as follows.

\begin{equation*}
\begin{aligned}
l_{n+1} &= l_n + \alpha * (N_{stm}^{rem} + N_{ltm}^{rem}) - N_{ltm}\\
&= l_n + \alpha * (N_{stm}^{rem} + N_{ltm}^{rem}) - l_n \times c\\
&= (1 - c) \times l_n + K \\
K &= \alpha * (N_{stm}^{rem} + N_{ltm}^{rem})\\
\end{aligned}
\end{equation*}

\begin{equation*}
\begin{aligned}
l_{n+1} - \frac{K}{c} &= (1-c) \times (l_n - \frac{K}{c})\\
b_{n+1} &= (1-c) \times b_n
\end{aligned}
\end{equation*}

\begin{equation*}
\begin{aligned}
b_n &= b_0 \times (1-c)^{n}\\
l_n &= b_0 \times (1-c)^n + \frac{K}{c}\\
&= b_0 \times (1-c)^n + \frac{\alpha * (N_{stm}^{rem} + N_{ltm}^{rem})}{c}
\end{aligned}
\end{equation*}

\begin{equation*}
\begin{aligned}
\lim_{n\to\infty} l_n &= \frac{\alpha * (N_{stm}^{rem} + N_{ltm}^{rem})}{c}\\
\lim_{n\to\infty} N_{ltm} &= \alpha * (N_{stm}^{rem} + N_{ltm}^{rem})\\
\end{aligned}
\end{equation*}

Ultimately, when a sufficient amount of time elapses, the overall sum of the lifespan will be proportionate to $\alpha * (N_{stm}^{rem} + N_{ltm}^{rem})$. Therefore, in the worst-case scenario of retrieve stage, the time complexity is as follows.

\begin{equation*}
\begin{aligned}
 O(N_{stm}^{rem} \times N_{ltm} \times N_{depth}) &= O(N_{stm}^{rem} \times (\alpha * (N_{stm}^{rem} + N_{ltm}^{rem})) \times N_{depth}) \\
 &= O(\alpha N_{stm}^{rem} N_{depth} (N_{stm}^{rem} + N_{ltm}^{rem}))
\end{aligned}
\end{equation*}

\begin{algorithm}
\begin{algorithmic}
    \caption{Exploit Stage}

    \STATE {\bfseries Input:} model $M$, input segment $s$, $retrieved$
    \STATE {\bfseries Output:} segment result $r$, hidden states $h_p$
    \STATE {\bfseries Result:} Conduct inference with retrieved memories. Return the segment result, hidden states, and attention weight for each engram.

    \STATE $r, h_p, a \gets M(s, retrieved)$
    \COMMENT{"a" means memory attention weights}
\end{algorithmic}
\end{algorithm}

The time complexity of the exploit stage depends upon the way of modeling utilization of retrieved engrams. In our implementation, we employed the cross-attention mechanism, wherein input data is used as a query for engrams serving as key and value. Consequently, the time complexity aligns with that of the cross-attention. The time complexity, given an input length of $L$ and the number of retrieved engrams $N_e$, is $O(L \times N_e)$. $N_e$ is equal to $N_{stm}^{rem} + N_{ltm}^{rem}$, so the time complexity is $O(L \times (N_{stm}^{rem} + N_{ltm}^{rem}))$. We configured the total number of engrams used in our experiments to be equal to the sequence length. In this scenario, the time complexity becomes $O(L^2)$, equivalent to that of the self-attention, thereby not exerting an additional impact on the overall time complexity from a Big-O perspective.

\begin{algorithm}
\begin{algorithmic}
    \caption{Memorize \& Forget Stage}

    \STATE {\bfseries Input:} $WM$, $STM$, $LTM$, $P$.
    \STATE {\bfseries Output:} Updated memories and condition tables.

    \STATE $P \gets AdjustConditionalProbs$($P$, $retrieved$)
    \COMMENT{update probs}
    \STATE $IncreaseLifespans$($retrieved$, $a$)
    \STATE $STM \gets MoveWMtoSTM$($WM$, $STM$)
    \STATE $DecreaseLifespanAndRemove$($STM$, $LTM$)
    \STATE $LTM \gets MoveSTMtoLTM$($STM$, $LTM$)
\end{algorithmic}
\end{algorithm}

The logic governing conditional probability adjustment increases the value for each pair of the retrieved engrams, resulting in a time complexity of $O(N_e^2)$. The logic regulating lifespan, being an operation for each engram, entails a complexity of $O(N_e)$. Changing the type of memory requires operations proportional to the number of engrams, limiting the complexity to $O(N_e)$. Consequently, the overall time complexity at this stage is $O(N_e^2) = O((N_{stm}^{rem} + N_{ltm}^{rem})^2)$.

In Memoria, space complexity is essentially the cost of maintaining a conditional probability table representing the connectivity between each engram. The space complexity is dependent on the implementation of the graph. For the sake of convenient implementation, we employed the adjacency matrix representation. When using an adjacency matrix, the spatial complexity becomes quadratic in the number of nodes, specifically, the square of the total number of engrams in Memoria, calculated as $O((N_{wm} + C_{stm} + N_{ltm})^2)$. Alternative implementations such as adjacency lists can further reduce spatial complexity.

\begin{table}[hb!]
\centering \small
\setlength{\tabcolsep}{3.5ex}
\caption{Time and space complexities on each stage.}
\begin{tabular}{lcc}
\toprule
Stage & Time Complexity & Space Complexity \\
\midrule
Retrieve & $O(N_{stm}^{rem} N_{ltm} N_{depth})$ & $O((N_{wm} + C_{stm} + N_{ltm})^2)$ \\
Exploit & $O(L (N_{stm}^{rem} + N_{ltm}^{rem}))$ & $O((N_{wm} + C_{stm} + N_{ltm})^2)$ \\
Memorize \& Forget & $O((N_{stm}^{rem} + N_{ltm}^{rem})^2)$ & \ $O((N_{wm} + C_{stm} + N_{ltm})^2)$ \\
\bottomrule
\end{tabular}
\vspace{0.1in}
\label{table:computational-complexities-summary}
\end{table}

\cref{table:computational-complexities-summary} shows the time complexity and space complexity for each stage using Big-O notation.

\subsection{Empirical Analysis}
\label{subsec:empirical-analysis}

\begin{table}[htb!]
\centering \small
\setlength{\tabcolsep}{3.5ex}
\caption{Inference time and GPU memory usage}
\begin{tabular}{lcc}
\toprule
Model & Execution Time (s) & Memory Usage (MB) \\
\midrule
Transformer-XL & 21.23 & 525.74 \\
Compressive Transformer & 25.57 & 740.00 \\
$\infty$-former & 54.15 & 676.12 \\
Memoria Transformer & 44.21 & 612.94 \\
\bottomrule
\end{tabular}
\vspace{0.1in}
\label{table:inference-time-and-memory-usage}
\end{table}

In addition to theoretical analysis, we measured the resources used during inference. \cref{table:inference-time-and-memory-usage} presents the time taken and average memory usage while inferring the Wikitext-103 test set. Regarding execution time, Memoria was faster than the relatively $\infty$-former but slower than the others, while its memory usage was the second smallest. Memory measurements were taken at each step using the functions \texttt{torch.cuda.empty\_cache} and \texttt{torch.cuda.memory\_allocated} provided by PyTorch, and averages were calculated.
In the case of Memoria, there is room for optimizing inference time and memory usage. This can be achieved by uploading information, such as engram type or lifespan, which requires minimal GPU computation, to the CPU. Additionally, optimization of data structures using adjacency lists can be applied. Furthermore, the graph maintained by Memoria does not necessarily need to be implemented using PyTorch or Python, allowing for increased computational efficiency through the use of more efficient programming languages.

%% file: pages/appendix/RAG-Discussions.tex
Recently, there has been considerable interest in Retrieval Augmented Generation (RAG) \citep{rag}, which has demonstrated remarkable performance in language modeling. Therefore, one might be inclined to compare Memoria with RAG. Can RAG serve as a substitute for Memoria?
The goal of our research is to address the problem of long-sequence processing. Language modeling is one of the tasks we performed to assess the capabilities of Memoria. Given that language modeling entails various intertwined challenges, there would be several ways to improve performance.
While RAG effectively enhances performance by providing external information as input to compensate for the limited internal knowledge of a language model, it is not directly associated with addressing the problem of processing long sequences.
Moreover, RAG employs retrieved text as additional input tokens, potentially requiring more long sequence processing capabilities. In such cases, we anticipate that Memoria could be utilized to effectively handle the increased sequence length resulting from the long retrieved texts.
Therefore, we compared Memoria with other conventional external memory approaches to validate its long sequence processing performance and did not conduct a comparison with RAG. Typical RAG and Memoria have different objectives aiming to address.

Could we then utilize the past input segments simply as a retrieval library for RAG, enabling its application in the context of long sequence processing? Unfortunately, when we consider applying the typical RAG technique to long sequence processing problems, there seem to be many difficulties.
Primarily, from a practical implementation perspective, maintaining the retrieval index poses a significant hurdle. When utilizing retrieval techniques without efficient indexing methods, a full search across the entire retrieval library becomes necessary. Hence, optimization techniques using tools such as Faiss \citep{faiss} or ScaNN \citep{scann} are commonly employed to construct retrieval indexes. While these indexing methods substantially reduce the inference time during retrieval, the initial indexing itself requires extensive time and computational resources. Efficient indexing techniques \citep{lsh, ivf, hnsw, scann} require a training phase to optimize inference according to the specific data distribution, causing the challenges when adding new items or removing outdated ones. Due to the nature of this task where new retrieval candidates are added with each inference, it is anticipated that utilizing indexing techniques would be hard, and substantial computational costs would be required to handle long sequences. Additionally, unlike memory-based methods, there is a problem of ever-increasing computational costs due to the absence of forgetting mechanisms.
Furthermore, considering the relatively short model input length in the segmentation and recursive inference mechanism, RAG would require a longer model input length to include enough text in the prompt to aid the model's inference, as the retrieved text occupies the input length. 

Apart from these issues, it seems essential to thoroughly explore aspects of performance such as how to compose the retrieval pool at the sentence level, paragraph level, or even at the entire segment level, how to train the retriever model, and how many retrievals should be conducted.
Additionally, since RAG is designed for text generation rather than general sequence processing, it cannot be simply applied to other tasks including sorting and classification and is limited to text generation only.
Certainly, if these constraints can be overcome, there is room for the emergence of a long-term memory technique based on RAG.
However, it is premature to discuss specific comparisons at this point.
If it happens, one potential benefit of RAGs, which are presently quite predictable, lies in the fact that their memory is stored in a text format that should be human-interpretable.

%% file: pages/appendix/Memoria-applied-transformers.tex
\subsection{Memoria Transformer}

\begin{figure}[h!]
    \centerline{\includegraphics[width=0.8\textwidth]{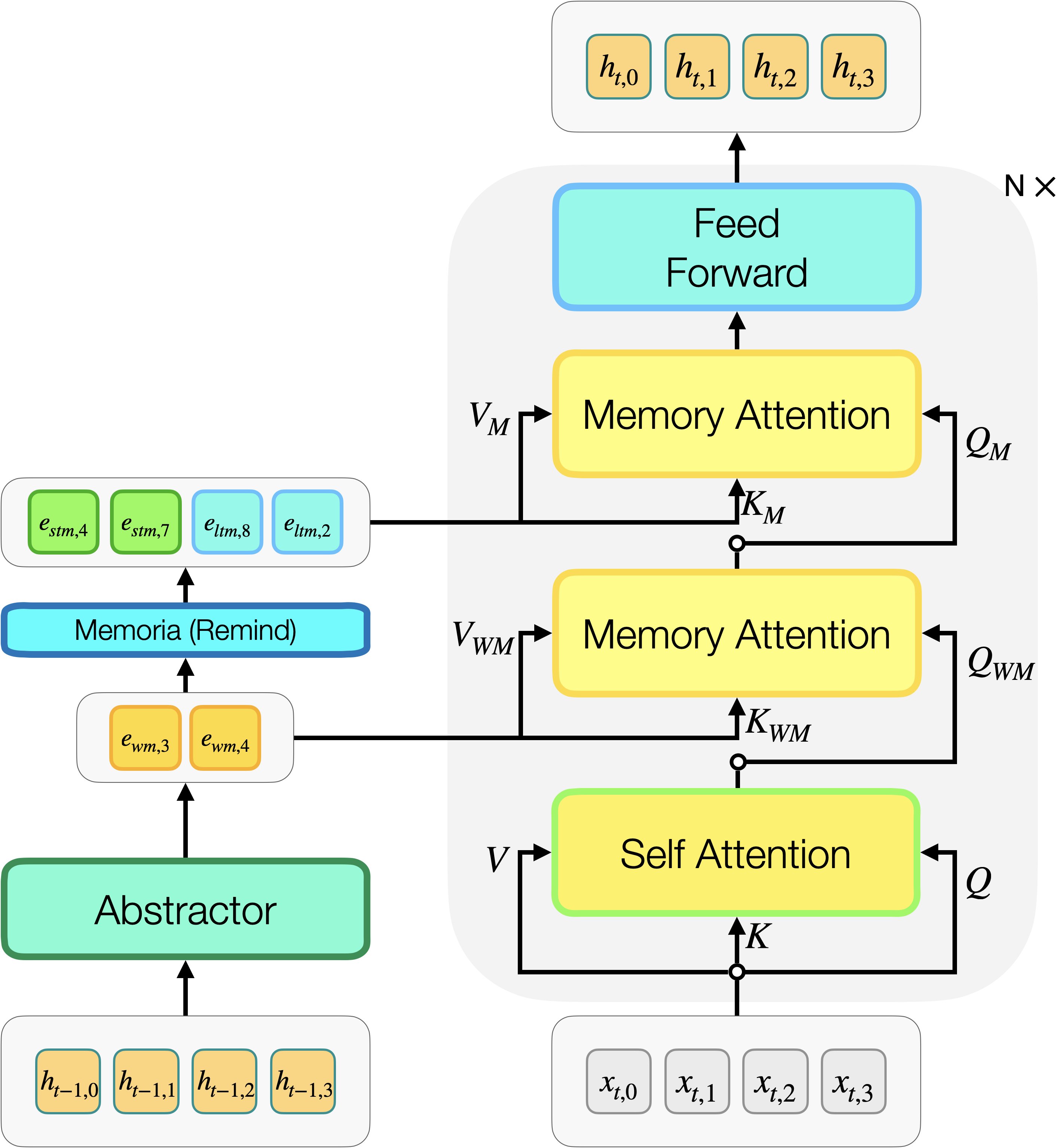}}
    \caption{Architecture of Memoria Transformer.
$t$ represents the current time step, and $x$ is the input embedding. The residual network and layer normalization are omitted for clarity. Memoria Transformer creates engrams from the previous time step output $h_{t-1}$ and retrieves engrams from short-term and long-term memory. Memoria Transformer exploits the engrams with the cross-attention. The memory attention blocks, depicted as two blocks in the diagram, are actually a single layer that shares the same weights.
}
\label{fig:memoria-gpt2}
\end{figure}

\newpage
\subsection{Memoria BERT/RoBERTa}

\begin{figure}[h!]
    \centerline{\includegraphics[width=0.8\textwidth]{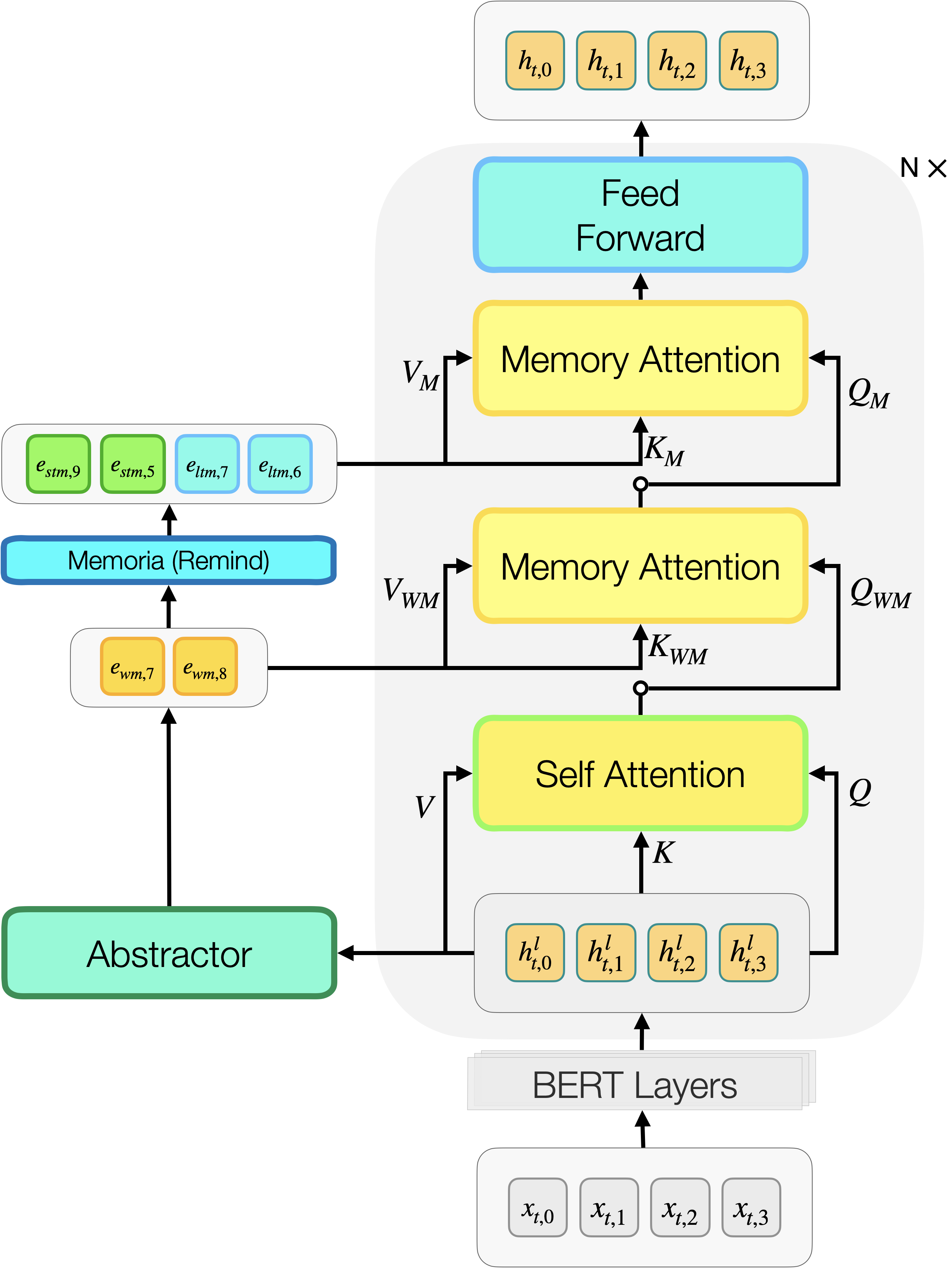}}
    \caption{Architecture of Memoria BERT/RoBERTa.
$t$ represents the current time step, and $x$ is the input embedding. The residual network and layer normalization are omitted for clarity. In the encoder models, unlike in the decoder model, engrams are created using information from the current time step. $l$ represents the memory layer index, and from layer 1 to layer $l$, each layer is a regular BERT layer. Using the output $h_t^l$ from layer $l$, engrams are created and retrieved from short-term and long-term memory. These engrams are then utilized in the subsequent layers (after layer $l$) through the cross-attention mechanism. The two memory attention blocks share the same weights, as in Memoria Transformer.
}
\label{fig:memoria-bert}
\end{figure}

%% file: pages/appendix/Visualization-Of-Memoria.tex
\begin{figure}[h!]
    \centering
    \begin{subfigure}[h!]{0.8\textwidth}
        \centering
        \includegraphics[width=0.55\textwidth]{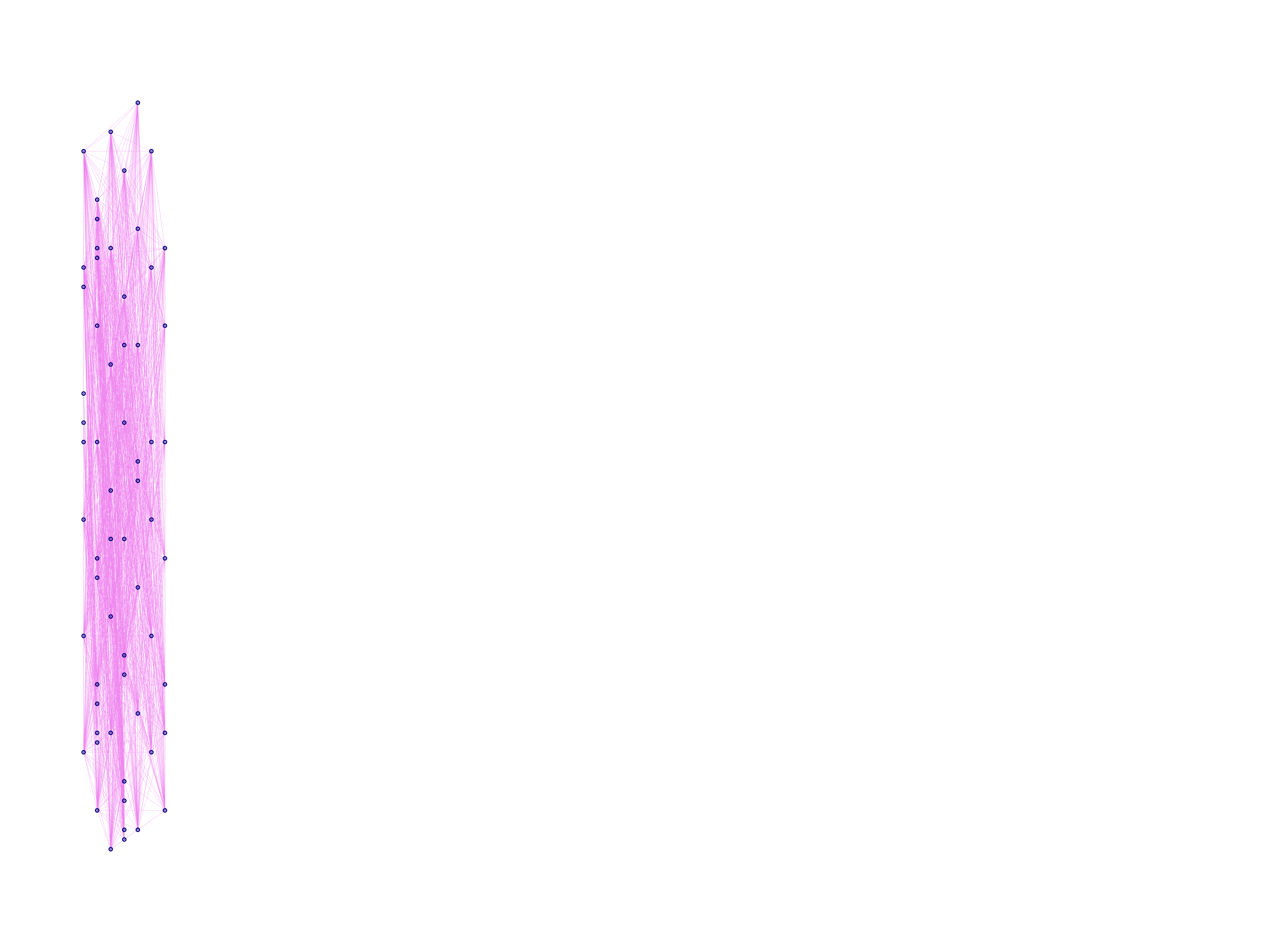}
        \caption{$t=20$}
    \end{subfigure}
    
    \begin{subfigure}[h!]{0.8\textwidth}
        \centering
        \includegraphics[width=0.55\textwidth]{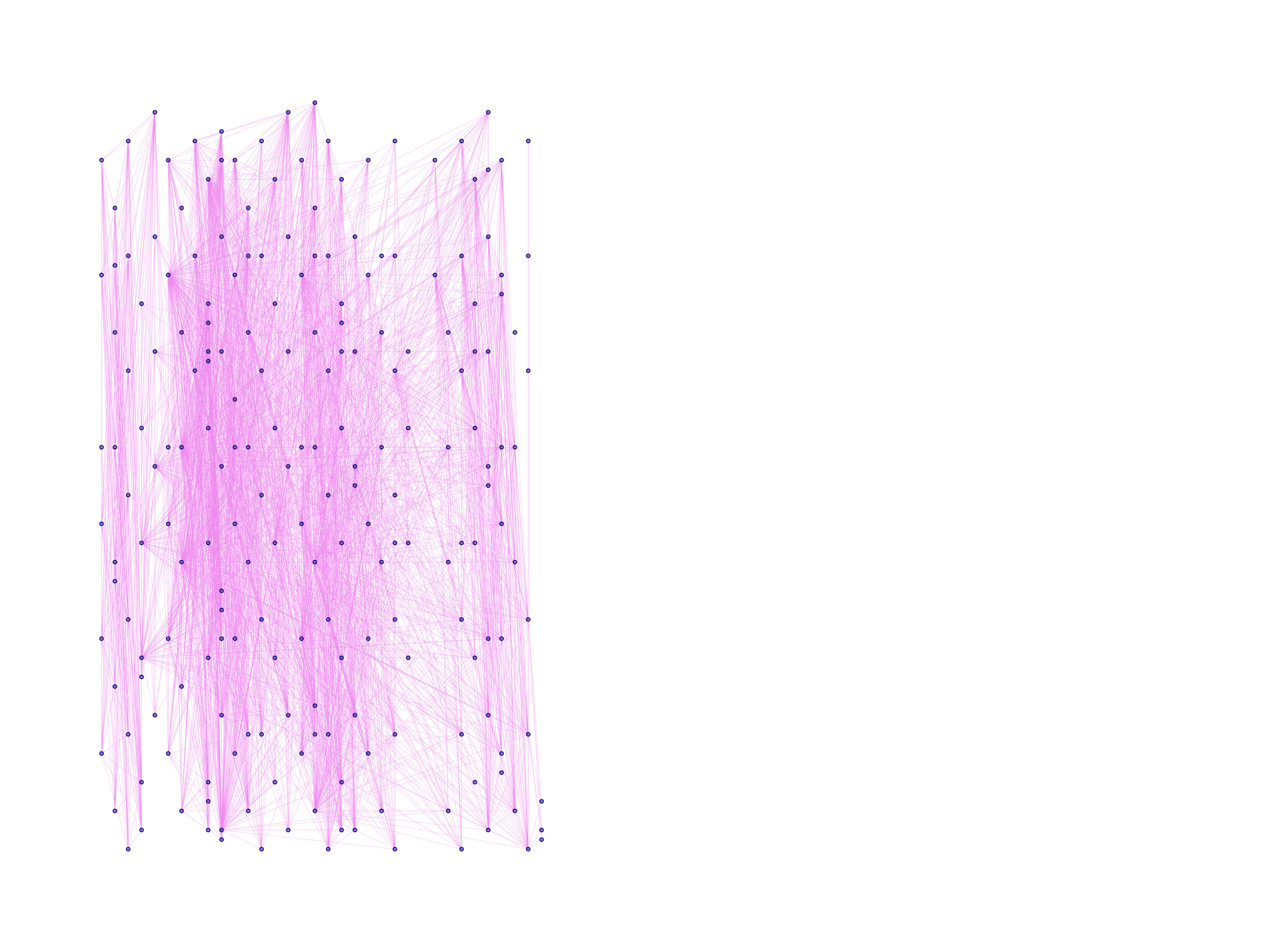}
        \caption{$t=60$}
    \end{subfigure}
    
    \begin{subfigure}[h!]{0.8\textwidth}
        \centering
        \includegraphics[width=0.55\textwidth]{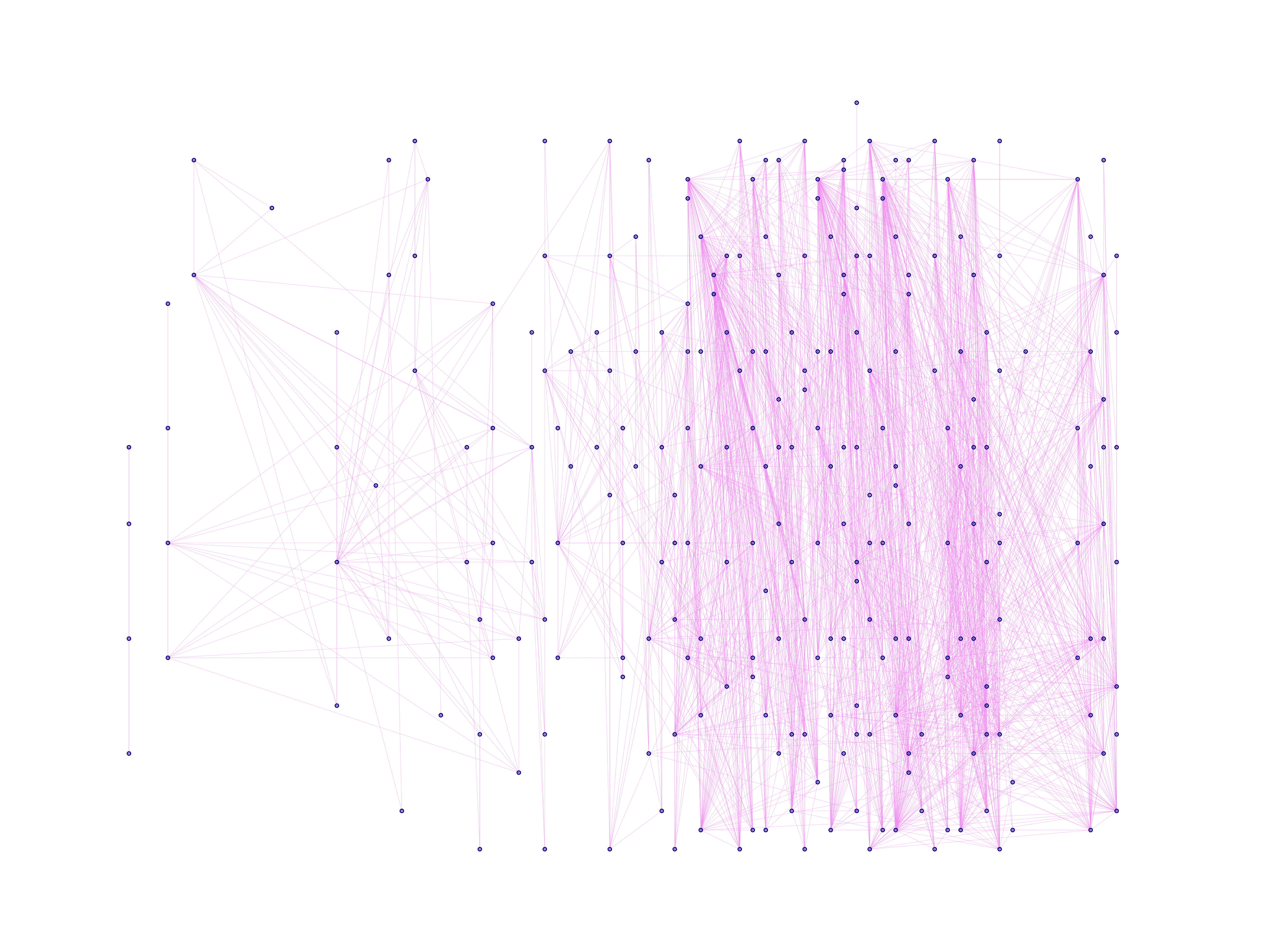}
        \caption{$t=130$}
    \end{subfigure}
     
    \caption{Changes in engrams of Memoria over time. The dots represent engrams, and the lines represent connections between engrams. $t$ is the time step. The farther to the right an engram is, the later it was created. Only the connections with high weights are shown for clarity. The engrams gradually fade away, but some important engrams still remain for a longer duration. This demonstrates Memoria's ability to preserve information, even if it has been a long time, as long as it remains useful. The strong nearby connections imply the humans' pattern of the temporal contiguity effect \citep{temporal-contiguity-effect}.}
\end{figure}

\begin{figure}[t!]
    \centering
    \begin{subfigure}[h!]{0.8\textwidth}
        \centering
        \includegraphics[width=0.55\textwidth]{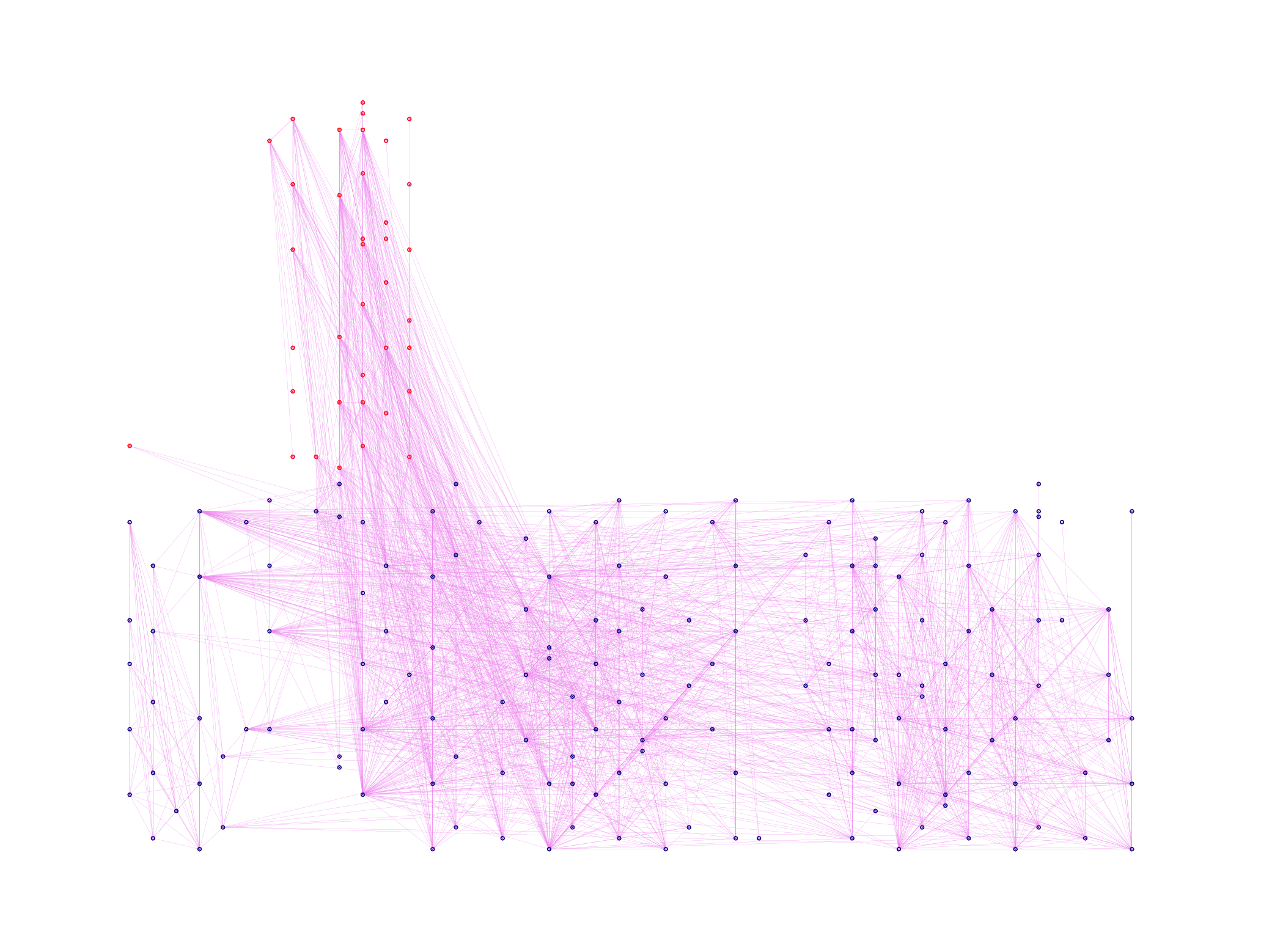}
        \caption{$t=87$}
    \end{subfigure}
    
    \begin{subfigure}[h!]{0.8\textwidth}
        \centering
        \includegraphics[width=0.55\textwidth]{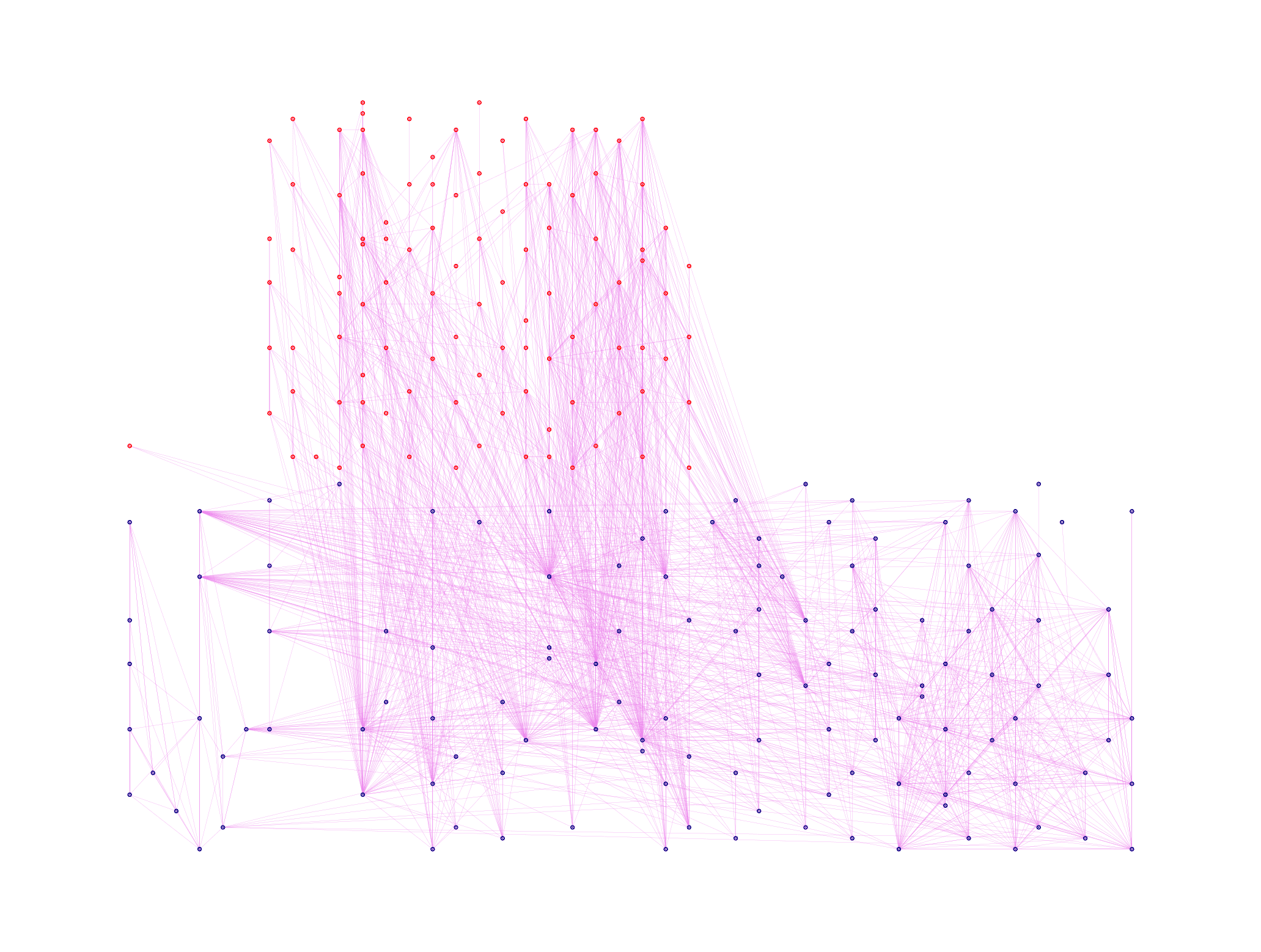}
        \caption{$t=105$}
    \end{subfigure}
    
    \begin{subfigure}[h!]{0.8\textwidth}
        \centering
        \includegraphics[width=0.55\textwidth]{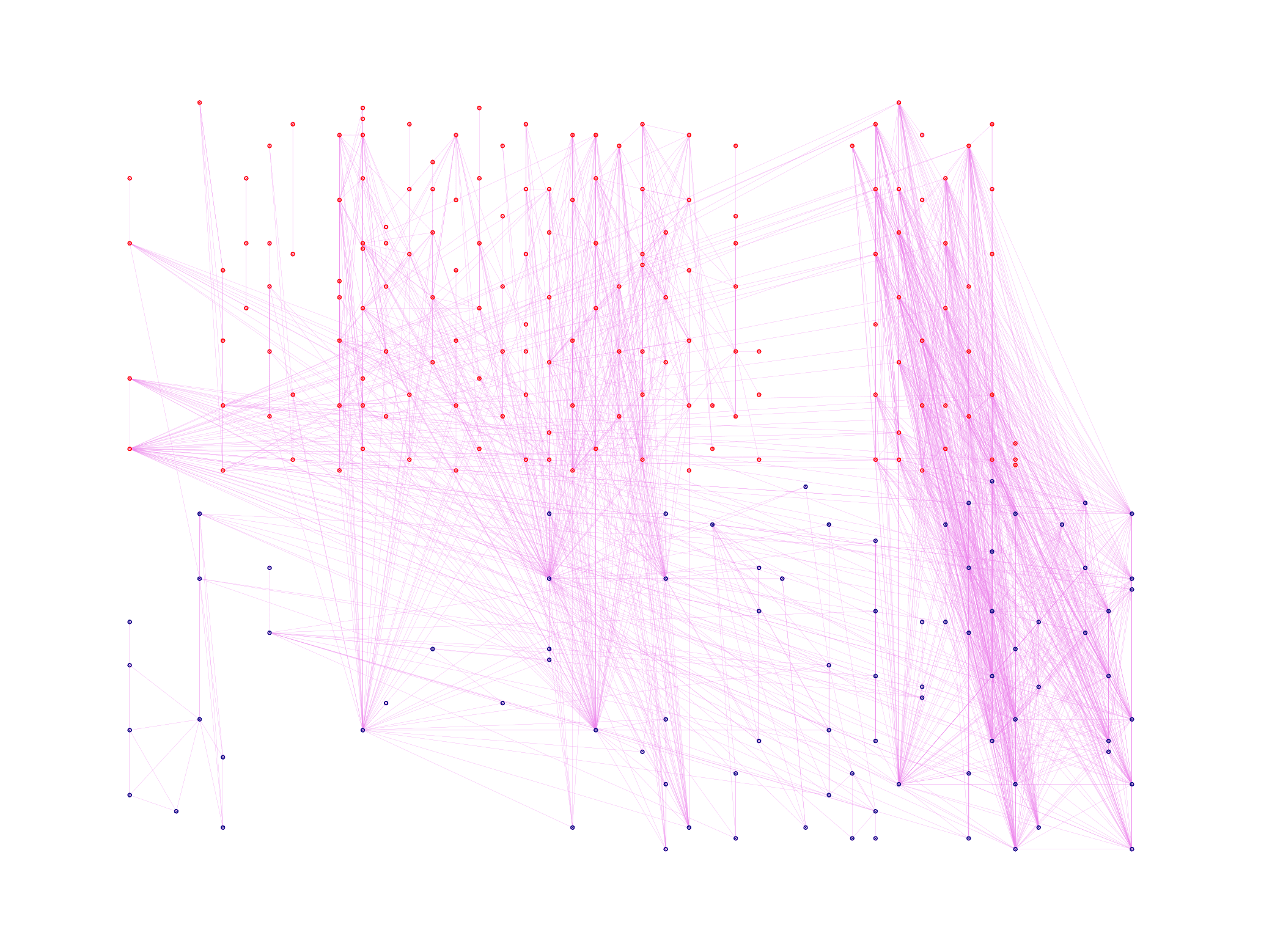}
        \caption{$t=122$}
    \end{subfigure}
     
    \caption{Changes in engrams of Memoria over time when Memoria sees the same data twice. The lower half of each image represents the engrams generated when observing at first, while the upper half represents the engrams generated when observing at second. Thus, the dots in the same vertical column represent engrams created from the same data. Engrams from the same data exhibit a generally stronger connectivity. This suggests that Memoria can consider content-based similarities between information even if they are temporally distant.}
\end{figure}

%% file: main.bbl
\begin{thebibliography}{76}
\providecommand{\natexlab}[1]{#1}
\providecommand{\url}[1]{\texttt{#1}}
\expandafter\ifx\csname urlstyle\endcsname\relax
  \providecommand{\doi}[1]{doi: #1}\else
  \providecommand{\doi}{doi: \begingroup \urlstyle{rm}\Url}\fi

\bibitem[Andoni et~al.(2015)Andoni, Indyk, Laarhoven, Razenshteyn, and Schmidt]{lsh}
Andoni, A., Indyk, P., Laarhoven, T., Razenshteyn, I., and Schmidt, L.
\newblock Practical and optimal lsh for angular distance.
\newblock In Cortes, C., Lawrence, N., Lee, D., Sugiyama, M., and Garnett, R. (eds.), \emph{Advances in Neural Information Processing Systems}, volume~28. Curran Associates, Inc., 2015.
\newblock URL \url{https://proceedings.neurips.cc/paper_files/paper/2015/file/2823f4797102ce1a1aec05359cc16dd9-Paper.pdf}.

\bibitem[Antony et~al.(2017)Antony, Ferreira, Norman, and Wimber]{retrieval-fast-route}
Antony, J.~W., Ferreira, C.~S., Norman, K.~A., and Wimber, M.
\newblock Retrieval as a fast route to memory consolidation.
\newblock \emph{Trends in cognitive sciences}, 21\penalty0 (8):\penalty0 573--576, 2017.

\bibitem[Atkinson \& Shiffrin(1968)Atkinson and Shiffrin]{multistore-model}
Atkinson, R. and Shiffrin, R.
\newblock Human memory: A proposed system and its control processes.
\newblock volume~2 of \emph{Psychology of Learning and Motivation}, pp.\  89--195. Academic Press, 1968.
\newblock \doi{https://doi.org/10.1016/S0079-7421(08)60422-3}.
\newblock URL \url{https://www.sciencedirect.com/science/article/pii/S0079742108604223}.

\bibitem[Atkinson \& Juola(1974)Atkinson and Juola]{search-decision-process}
Atkinson, R.~C. and Juola, J.~F.
\newblock \emph{Search and decision processes in recognition memory.}, pp.\  xiii, 299--xiii, 299.
\newblock Contemporary developments in mathematical psychology: I. Learning, memory and thinking. W. H. Freeman, Oxford, England, 1974.
\newblock ISBN 0716708485.

\bibitem[Atkinson et~al.(1974)Atkinson, Hermann, and Wescourt]{search-process-recognition}
Atkinson, R.~C., Hermann, D.~J., and Wescourt, K.~T.
\newblock \emph{Search processes in recognition memory.}, pp.\  xi, 386--xi, 386.
\newblock Theories in cognitive psychology: The Loyola Symposium. Lawrence Erlbaum, Oxford, England, 1974.
\newblock ISBN 0470812281.

\bibitem[Baddeley \& Hitch(1993)Baddeley and Hitch]{recency-effect2}
Baddeley, A.~D. and Hitch, G.
\newblock The recency effect: Implicit learning with explicit retrieval?
\newblock \emph{Memory \& Cognition}, 21\penalty0 (2):\penalty0 146--155, 1993.
\newblock \doi{10.3758/BF03202726}.
\newblock URL \url{https://doi.org/10.3758/BF03202726}.

\bibitem[Banino et~al.(2020)Banino, Badia, Köster, Chadwick, Zambaldi, Hassabis, Barry, Botvinick, Kumaran, and Blundell]{memo}
Banino, A., Badia, A.~P., Köster, R., Chadwick, M.~J., Zambaldi, V., Hassabis, D., Barry, C., Botvinick, M., Kumaran, D., and Blundell, C.
\newblock Memo: A deep network for flexible combination of episodic memories.
\newblock In \emph{International Conference on Learning Representations}, 2020.
\newblock URL \url{https://openreview.net/forum?id=rJxlc0EtDr}.

\bibitem[Baranchuk et~al.(2018)Baranchuk, Babenko, and Malkov]{ivf}
Baranchuk, D., Babenko, A., and Malkov, Y.
\newblock Revisiting the inverted indices for billion-scale approximate nearest neighbors.
\newblock In Ferrari, V., Hebert, M., Sminchisescu, C., and Weiss, Y. (eds.), \emph{Computer Vision -- ECCV 2018}, pp.\  209--224, Cham, 2018. Springer International Publishing.
\newblock ISBN 978-3-030-01258-8.

\bibitem[Beltagy et~al.(2020)Beltagy, Peters, and Cohan]{longformer}
Beltagy, I., Peters, M.~E., and Cohan, A.
\newblock Longformer: The long-document transformer.
\newblock \emph{CoRR}, abs/2004.05150, 2020.
\newblock URL \url{https://arxiv.org/abs/2004.05150}.

\bibitem[Bjork(1975)]{retrieval-as-memory-modifier}
Bjork, R.
\newblock Retrieval as a memory modifier: An interpretation of negative recency \& related phenomena.
\newblock \emph{Information Processing and Cognition: The Loyola Symposium}, 01 1975.

\bibitem[Bjork(1988)]{retrieval-practice}
Bjork, R.~A.
\newblock \emph{Retrieval practice and the maintenance of knowledge.}, pp.\  396--401.
\newblock Practical aspects of memory: Current research and issues, Vol. 1: Memory in everyday life. John Wiley {\&} Sons, Oxford, England, 1988.
\newblock ISBN 0-471-91234-4 (Hardcover).

\bibitem[Bjork \& Whitten(1974)Bjork and Whitten]{recency-effect}
Bjork, R.~A. and Whitten, W.~B.
\newblock Recency-sensitive retrieval processes in long-term free recall.
\newblock \emph{Cognitive Psychology}, 6\penalty0 (2):\penalty0 173--189, 1974.
\newblock ISSN 0010-0285.
\newblock \doi{https://doi.org/10.1016/0010-0285(74)90009-7}.
\newblock URL \url{https://www.sciencedirect.com/science/article/pii/0010028574900097}.

\bibitem[Braun et~al.(2018)Braun, Wimmer, and Shohamy]{brain-keep-high-reward-memory}
Braun, E.~K., Wimmer, G.~E., and Shohamy, D.
\newblock Retroactive and graded prioritization of memory by reward.
\newblock \emph{Nature Communications}, 9\penalty0 (1):\penalty0 4886, 2018.
\newblock \doi{10.1038/s41467-018-07280-0}.
\newblock URL \url{https://doi.org/10.1038/s41467-018-07280-0}.

\bibitem[Brown(1958)]{trace-decay-brown}
Brown, J.
\newblock Some tests of the decay theory of immediate memory.
\newblock \emph{Quarterly Journal of Experimental Psychology}, 10\penalty0 (1):\penalty0 12--21, 1958.
\newblock \doi{10.1080/17470215808416249}.
\newblock URL \url{https://doi.org/10.1080/17470215808416249}.

\bibitem[Brown et~al.(2020)Brown, Mann, Ryder, Subbiah, Kaplan, Dhariwal, Neelakantan, Shyam, Sastry, Askell, Agarwal, Herbert-Voss, Krueger, Henighan, Child, Ramesh, Ziegler, Wu, Winter, Hesse, Chen, Sigler, Litwin, Gray, Chess, Clark, Berner, McCandlish, Radford, Sutskever, and Amodei]{gpt}
Brown, T., Mann, B., Ryder, N., Subbiah, M., Kaplan, J.~D., Dhariwal, P., Neelakantan, A., Shyam, P., Sastry, G., Askell, A., Agarwal, S., Herbert-Voss, A., Krueger, G., Henighan, T., Child, R., Ramesh, A., Ziegler, D., Wu, J., Winter, C., Hesse, C., Chen, M., Sigler, E., Litwin, M., Gray, S., Chess, B., Clark, J., Berner, C., McCandlish, S., Radford, A., Sutskever, I., and Amodei, D.
\newblock Language models are few-shot learners.
\newblock In Larochelle, H., Ranzato, M., Hadsell, R., Balcan, M., and Lin, H. (eds.), \emph{Advances in Neural Information Processing Systems}, volume~33, pp.\  1877--1901. Curran Associates, Inc., 2020.
\newblock URL \url{https://proceedings.neurips.cc/paper_files/paper/2020/file/1457c0d6bfcb4967418bfb8ac142f64a-Paper.pdf}.

\bibitem[Bulatov et~al.(2022)Bulatov, Kuratov, and Burtsev]{rmt}
Bulatov, A., Kuratov, Y., and Burtsev, M.
\newblock Recurrent memory transformer.
\newblock In Koyejo, S., Mohamed, S., Agarwal, A., Belgrave, D., Cho, K., and Oh, A. (eds.), \emph{Advances in Neural Information Processing Systems}, volume~35, pp.\  11079--11091. Curran Associates, Inc., 2022.
\newblock URL \url{https://proceedings.neurips.cc/paper_files/paper/2022/file/47e288629a6996a17ce50b90a056a0e1-Paper-Conference.pdf}.

\bibitem[Burtsev \& Sapunov(2020)Burtsev and Sapunov]{memory-transformer}
Burtsev, M.~S. and Sapunov, G.~V.
\newblock Memory transformer.
\newblock \emph{CoRR}, abs/2006.11527, 2020.
\newblock URL \url{https://arxiv.org/abs/2006.11527}.

\bibitem[Caporale \& Dan(2008)Caporale and Dan]{hebbian-learning-rule}
Caporale, N. and Dan, Y.
\newblock Spike timing-dependent plasticity: a hebbian learning rule.
\newblock \emph{Annu Rev Neurosci}, 31:\penalty0 25--46, 2008.

\bibitem[Chung et~al.(2014)Chung, G{\"{u}}l{\c{c}}ehre, Cho, and Bengio]{gru}
Chung, J., G{\"{u}}l{\c{c}}ehre, {\c{C}}., Cho, K., and Bengio, Y.
\newblock Empirical evaluation of gated recurrent neural networks on sequence modeling.
\newblock \emph{CoRR}, abs/1412.3555, 2014.
\newblock URL \url{http://arxiv.org/abs/1412.3555}.

\bibitem[Craik \& Lockhart(1972)Craik and Lockhart]{levels-of-processing}
Craik, F.~I. and Lockhart, R.~S.
\newblock Levels of processing: A framework for memory research.
\newblock \emph{Journal of Verbal Learning and Verbal Behavior}, 11\penalty0 (6):\penalty0 671--684, 1972.
\newblock ISSN 0022-5371.
\newblock \doi{https://doi.org/10.1016/S0022-5371(72)80001-X}.
\newblock URL \url{https://www.sciencedirect.com/science/article/pii/S002253717280001X}.

\bibitem[Dai et~al.(2019)Dai, Yang, Yang, Carbonell, Le, and Salakhutdinov]{transfoxl}
Dai, Z., Yang, Z., Yang, Y., Carbonell, J., Le, Q.~V., and Salakhutdinov, R.
\newblock Transformer-xl: Attentive language models beyond a fixed-length context.
\newblock \emph{ACL 2019 - 57th Annual Meeting of the Association for Computational Linguistics, Proceedings of the Conference}, pp.\  2978--2988, 1 2019.
\newblock \doi{10.48550/arxiv.1901.02860}.
\newblock URL \url{https://arxiv.org/abs/1901.02860v3}.

\bibitem[Davis et~al.(2014)Davis, Xue, Love, Preston, and Poldrack]{global-neural-similarity}
Davis, T., Xue, G., Love, B.~C., Preston, A.~R., and Poldrack, R.~A.
\newblock Global neural pattern similarity as a common basis for categorization and recognition memory.
\newblock \emph{Journal of Neuroscience}, 34\penalty0 (22):\penalty0 7472--7484, 2014.
\newblock ISSN 0270-6474.
\newblock \doi{10.1523/JNEUROSCI.3376-13.2014}.
\newblock URL \url{https://www.jneurosci.org/content/34/22/7472}.

\bibitem[Deese \& Kaufman(1957)Deese and Kaufman]{serial-position}
Deese, J. and Kaufman, R.~A.
\newblock Serial effects in recall of unorganized and sequentially organized verbal material.
\newblock \emph{Journal of Experimental Psychology}, 54\penalty0 (3):\penalty0 180--187, 1957.
\newblock ISSN 0022-1015(Print).
\newblock \doi{10.1037/h0040536}.
\newblock URL \url{https://doi.org/10.1037/h0040536}.

\bibitem[Devlin et~al.(2019)Devlin, Chang, Lee, and Toutanova]{bert}
Devlin, J., Chang, M.-W., Lee, K., and Toutanova, K.
\newblock {BERT}: Pre-training of deep bidirectional transformers for language understanding.
\newblock In \emph{Proceedings of the 2019 Conference of the North {A}merican Chapter of the Association for Computational Linguistics: Human Language Technologies, Volume 1 (Long and Short Papers)}, pp.\  4171--4186, Minneapolis, Minnesota, June 2019. Association for Computational Linguistics.
\newblock \doi{10.18653/v1/N19-1423}.
\newblock URL \url{https://aclanthology.org/N19-1423}.

\bibitem[Douze et~al.(2024)Douze, Guzhva, Deng, Johnson, Szilvasy, Mazaré, Lomeli, Hosseini, and Jégou]{faiss}
Douze, M., Guzhva, A., Deng, C., Johnson, J., Szilvasy, G., Mazaré, P.-E., Lomeli, M., Hosseini, L., and Jégou, H.
\newblock The faiss library.
\newblock 2024.

\bibitem[Gerstner \& Kistler(2002)Gerstner and Kistler]{math-hebb}
Gerstner, W. and Kistler, W.~M.
\newblock Mathematical formulations of hebbian learning.
\newblock \emph{Biological Cybernetics}, 87\penalty0 (5):\penalty0 404--415, 2002.
\newblock \doi{10.1007/s00422-002-0353-y}.
\newblock URL \url{https://doi.org/10.1007/s00422-002-0353-y}.

\bibitem[Ginns(2006)]{temporal-contiguity-effect2}
Ginns, P.
\newblock Integrating information: A meta-analysis of the spatial contiguity and temporal contiguity effects.
\newblock \emph{Learning and Instruction}, 16\penalty0 (6):\penalty0 511--525, 2006.
\newblock ISSN 0959-4752.
\newblock \doi{https://doi.org/10.1016/j.learninstruc.2006.10.001}.
\newblock URL \url{https://www.sciencedirect.com/science/article/pii/S0959475206000806}.

\bibitem[Graves et~al.(2014)Graves, Wayne, and Danihelka]{ntm}
Graves, A., Wayne, G., and Danihelka, I.
\newblock Neural turing machines.
\newblock \emph{CoRR}, abs/1410.5401, 2014.
\newblock URL \url{http://arxiv.org/abs/1410.5401}.

\bibitem[Graves et~al.(2016)Graves, Wayne, Reynolds, Harley, Danihelka, Grabska-Barwińska, Colmenarejo, Grefenstette, Ramalho, Agapiou, Badia, Hermann, Zwols, Ostrovski, Cain, King, Summerfield, Blunsom, Kavukcuoglu, and Hassabis]{dnc}
Graves, A., Wayne, G., Reynolds, M., Harley, T., Danihelka, I., Grabska-Barwińska, A., Colmenarejo, S.~G., Grefenstette, E., Ramalho, T., Agapiou, J., Badia, A.~P., Hermann, K.~M., Zwols, Y., Ostrovski, G., Cain, A., King, H., Summerfield, C., Blunsom, P., Kavukcuoglu, K., and Hassabis, D.
\newblock Hybrid computing using a neural network with dynamic external memory.
\newblock \emph{Nature}, 538\penalty0 (7626):\penalty0 471--476, October 2016.
\newblock ISSN 00280836.
\newblock URL \url{http://dx.doi.org/10.1038/nature20101}.

\bibitem[Gulcehre et~al.(2017{\natexlab{a}})Gulcehre, Chandar, and Bengio]{tardis}
Gulcehre, C., Chandar, S., and Bengio, Y.
\newblock Memory augmented neural networks with wormhole connections, 2017{\natexlab{a}}.

\bibitem[Gulcehre et~al.(2017{\natexlab{b}})Gulcehre, Chandar, Cho, and Bengio]{d-ntm}
Gulcehre, C., Chandar, S., Cho, K., and Bengio, Y.
\newblock Dynamic neural turing machine with soft and hard addressing schemes, 2017{\natexlab{b}}.

\bibitem[Guo et~al.(2020{\natexlab{a}})Guo, Sun, Lindgren, Geng, Simcha, Chern, and Kumar]{hnsw}
Guo, R., Sun, P., Lindgren, E., Geng, Q., Simcha, D., Chern, F., and Kumar, S.
\newblock Accelerating large-scale inference with anisotropic vector quantization.
\newblock In III, H.~D. and Singh, A. (eds.), \emph{Proceedings of the 37th International Conference on Machine Learning}, volume 119 of \emph{Proceedings of Machine Learning Research}, pp.\  3887--3896. PMLR, 13--18 Jul 2020{\natexlab{a}}.
\newblock URL \url{https://proceedings.mlr.press/v119/guo20h.html}.

\bibitem[Guo et~al.(2020{\natexlab{b}})Guo, Sun, Lindgren, Geng, Simcha, Chern, and Kumar]{scann}
Guo, R., Sun, P., Lindgren, E., Geng, Q., Simcha, D., Chern, F., and Kumar, S.
\newblock Accelerating large-scale inference with anisotropic vector quantization.
\newblock In III, H.~D. and Singh, A. (eds.), \emph{Proceedings of the 37th International Conference on Machine Learning}, volume 119 of \emph{Proceedings of Machine Learning Research}, pp.\  3887--3896. PMLR, 13--18 Jul 2020{\natexlab{b}}.
\newblock URL \url{https://proceedings.mlr.press/v119/guo20h.html}.

\bibitem[Hassabis et~al.(2017)Hassabis, Kumaran, Summerfield, and Botvinick]{neuro-inspired-ai}
Hassabis, D., Kumaran, D., Summerfield, C., and Botvinick, M.
\newblock Neuroscience-inspired artificial intelligence.
\newblock \emph{Neuron}, 95\penalty0 (2):\penalty0 245--258, 2017.
\newblock ISSN 0896-6273.
\newblock \doi{https://doi.org/10.1016/j.neuron.2017.06.011}.
\newblock URL \url{https://www.sciencedirect.com/science/article/pii/S0896627317305093}.

\bibitem[Hebb(1949)]{Hebb1949}
Hebb, D.~O.
\newblock \emph{The organization of behavior}.
\newblock 1949.

\bibitem[Hochreiter \& Schmidhuber(1997)Hochreiter and Schmidhuber]{lstm}
Hochreiter, S. and Schmidhuber, J.
\newblock Long short-term memory.
\newblock \emph{Neural Computation}, 9:\penalty0 1735--1780, 11 1997.
\newblock ISSN 08997667.
\newblock \doi{10.1162/NECO.1997.9.8.1735}.
\newblock URL \url{https://www.researchgate.net/publication/13853244_Long_Short-term_Memory}.

\bibitem[Journ{\'e} et~al.(2023)Journ{\'e}, Rodriguez, Guo, and Moraitis]{hebb-without-feedback}
Journ{\'e}, A., Rodriguez, H.~G., Guo, Q., and Moraitis, T.
\newblock Hebbian deep learning without feedback.
\newblock In \emph{The Eleventh International Conference on Learning Representations}, 2023.
\newblock URL \url{https://openreview.net/forum?id=8gd4M-_Rj1}.

\bibitem[Kahana(1996)]{temporal-contiguity-effect}
Kahana, M.~J.
\newblock Associative retrieval processes in free recall.
\newblock \emph{Memory \& Cognition}, 24\penalty0 (1):\penalty0 103--109, 1996.
\newblock \doi{10.3758/BF03197276}.
\newblock URL \url{https://doi.org/10.3758/BF03197276}.

\bibitem[Kahana(2020)]{compu-model-memory-search}
Kahana, M.~J.
\newblock Computational models of memory search.
\newblock \emph{Annual Review of Psychology}, 71\penalty0 (1):\penalty0 107--138, 2020.
\newblock \doi{10.1146/annurev-psych-010418-103358}.
\newblock URL \url{https://doi.org/10.1146/annurev-psych-010418-103358}.
\newblock PMID: 31567043.

\bibitem[Kiesel et~al.(2019)Kiesel, Mestre, Shukla, Vincent, Adineh, Corney, Stein, and Potthast]{hyperpartisan}
Kiesel, J., Mestre, M., Shukla, R., Vincent, E., Adineh, P., Corney, D., Stein, B., and Potthast, M.
\newblock {S}em{E}val-2019 task 4: Hyperpartisan news detection.
\newblock In \emph{Proceedings of the 13th International Workshop on Semantic Evaluation}, pp.\  829--839, Minneapolis, Minnesota, USA, June 2019. Association for Computational Linguistics.
\newblock \doi{10.18653/v1/S19-2145}.
\newblock URL \url{https://aclanthology.org/S19-2145}.

\bibitem[Kim et~al.(2023)Kim, Cochez, Fran\c{c}ois-Lavet, Neerincx, and Vossen]{machine-memory-system}
Kim, T., Cochez, M., Fran\c{c}ois-Lavet, V., Neerincx, M., and Vossen, P.
\newblock A machine with short-term, episodic, and semantic memory systems.
\newblock In \emph{Proceedings of the Thirty-Seventh AAAI Conference on Artificial Intelligence and Thirty-Fifth Conference on Innovative Applications of Artificial Intelligence and Thirteenth Symposium on Educational Advances in Artificial Intelligence}, AAAI'23/IAAI'23/EAAI'23. AAAI Press, 2023.
\newblock ISBN 978-1-57735-880-0.
\newblock \doi{10.1609/aaai.v37i1.25075}.
\newblock URL \url{https://doi.org/10.1609/aaai.v37i1.25075}.

\bibitem[Kitaev et~al.(2020)Kitaev, Kaiser, and Levskaya]{reformer}
Kitaev, N., Kaiser, L., and Levskaya, A.
\newblock Reformer: The efficient transformer.
\newblock \emph{CoRR}, abs/2001.04451, 2020.
\newblock URL \url{https://arxiv.org/abs/2001.04451}.

\bibitem[Kuriscak et~al.(2015)Kuriscak, Marsalek, Stroffek, and Toth]{hebb-nn}
Kuriscak, E., Marsalek, P., Stroffek, J., and Toth, P.~G.
\newblock Biological context of hebb learning in artificial neural networks, a review.
\newblock \emph{Neurocomputing}, 152:\penalty0 27--35, 2015.
\newblock ISSN 0925-2312.
\newblock \doi{https://doi.org/10.1016/j.neucom.2014.11.022}.
\newblock URL \url{https://www.sciencedirect.com/science/article/pii/S0925231214015239}.

\bibitem[Le et~al.(2020)Le, Tran, and Venkatesh]{self-attentive-asso}
Le, H., Tran, T., and Venkatesh, S.
\newblock Self-attentive associative memory.
\newblock In III, H.~D. and Singh, A. (eds.), \emph{Proceedings of the 37th International Conference on Machine Learning}, volume 119 of \emph{Proceedings of Machine Learning Research}, pp.\  5682--5691. PMLR, 13--18 Jul 2020.
\newblock URL \url{https://proceedings.mlr.press/v119/le20b.html}.

\bibitem[Lewis et~al.(2020{\natexlab{a}})Lewis, Liu, Goyal, Ghazvininejad, Mohamed, Levy, Stoyanov, and Zettlemoyer]{bart}
Lewis, M., Liu, Y., Goyal, N., Ghazvininejad, M., Mohamed, A., Levy, O., Stoyanov, V., and Zettlemoyer, L.
\newblock {BART}: Denoising sequence-to-sequence pre-training for natural language generation, translation, and comprehension.
\newblock In \emph{Proceedings of the 58th Annual Meeting of the Association for Computational Linguistics}, pp.\  7871--7880, Online, July 2020{\natexlab{a}}. Association for Computational Linguistics.
\newblock \doi{10.18653/v1/2020.acl-main.703}.
\newblock URL \url{https://aclanthology.org/2020.acl-main.703}.

\bibitem[Lewis et~al.(2020{\natexlab{b}})Lewis, Perez, Piktus, Petroni, Karpukhin, Goyal, K\"{u}ttler, Lewis, Yih, Rockt\"{a}schel, Riedel, and Kiela]{rag}
Lewis, P., Perez, E., Piktus, A., Petroni, F., Karpukhin, V., Goyal, N., K\"{u}ttler, H., Lewis, M., Yih, W.-t., Rockt\"{a}schel, T., Riedel, S., and Kiela, D.
\newblock Retrieval-augmented generation for knowledge-intensive nlp tasks.
\newblock In Larochelle, H., Ranzato, M., Hadsell, R., Balcan, M., and Lin, H. (eds.), \emph{Advances in Neural Information Processing Systems}, volume~33, pp.\  9459--9474. Curran Associates, Inc., 2020{\natexlab{b}}.
\newblock URL \url{https://proceedings.neurips.cc/paper_files/paper/2020/file/6b493230205f780e1bc26945df7481e5-Paper.pdf}.

\bibitem[Limbacher \& Legenstein(2020)Limbacher and Legenstein]{h-mem}
Limbacher, T. and Legenstein, R.
\newblock H-mem: Harnessing synaptic plasticity with hebbian memory networks.
\newblock In Larochelle, H., Ranzato, M., Hadsell, R., Balcan, M., and Lin, H. (eds.), \emph{Advances in Neural Information Processing Systems}, volume~33, pp.\  21627--21637. Curran Associates, Inc., 2020.
\newblock URL \url{https://proceedings.neurips.cc/paper_files/paper/2020/file/f6876a9f998f6472cc26708e27444456-Paper.pdf}.

\bibitem[Mahoney(2006)]{enwik8}
Mahoney, M.
\newblock Large text compression benchmark, 2006.
\newblock URL \url{http://www.mattmahoney.net/dc/text.html}.

\bibitem[Martins et~al.(2021)Martins, Marinho, and Martins]{infformer}
Martins, P.~H., Marinho, Z., and Martins, A. F.~T.
\newblock $\infty$-former: Infinite memory transformer.
\newblock 9 2021.
\newblock \doi{10.48550/arxiv.2109.00301}.
\newblock URL \url{https://arxiv.org/abs/2109.00301v3}.

\bibitem[Meng \& Rumshisky(2018)Meng and Rumshisky]{gcl}
Meng, Y. and Rumshisky, A.
\newblock Context-aware neural model for temporal information extraction.
\newblock In Gurevych, I. and Miyao, Y. (eds.), \emph{Proceedings of the 56th Annual Meeting of the Association for Computational Linguistics (Volume 1: Long Papers)}, pp.\  527--536, Melbourne, Australia, July 2018. Association for Computational Linguistics.
\newblock \doi{10.18653/v1/P18-1049}.
\newblock URL \url{https://aclanthology.org/P18-1049}.

\bibitem[Merity et~al.(2017)Merity, Xiong, Bradbury, and Socher]{wikitext103}
Merity, S., Xiong, C., Bradbury, J., and Socher, R.
\newblock Pointer sentinel mixture models.
\newblock In \emph{5th International Conference on Learning Representations, {ICLR} 2017, Toulon, France, April 24-26, 2017, Conference Track Proceedings}. OpenReview.net, 2017.
\newblock URL \url{https://openreview.net/forum?id=Byj72udxe}.

\bibitem[Morrissey et~al.(2017)Morrissey, Insel, and Takehara-Nishiuchi]{how-brain-maintain-useful-memory}
Morrissey, M.~D., Insel, N., and Takehara-Nishiuchi, K.
\newblock Generalizable knowledge outweighs incidental details in prefrontal ensemble code over time.
\newblock \emph{eLife}, 6:\penalty0 e22177, feb 2017.
\newblock ISSN 2050-084X.
\newblock \doi{10.7554/eLife.22177}.
\newblock URL \url{https://doi.org/10.7554/eLife.22177}.

\bibitem[Murdock~Jr.(1962)]{serial-position2}
Murdock~Jr., B.~B.
\newblock The serial position effect of free recall.
\newblock \emph{Journal of Experimental Psychology}, 64\penalty0 (5):\penalty0 482--488, 1962.
\newblock ISSN 0022-1015(Print).
\newblock \doi{10.1037/h0045106}.
\newblock URL \url{https://doi.org/10.1037/h0045106}.

\bibitem[Nairne \& Pandeirada(2008)Nairne and Pandeirada]{adaptive-memory}
Nairne, J.~S. and Pandeirada, J.~N.
\newblock Adaptive memory: Remembering with a stone-age brain.
\newblock \emph{Current Directions in Psychological Science}, 17\penalty0 (4):\penalty0 239--243, 2008.
\newblock \doi{10.1111/j.1467-8721.2008.00582.x}.
\newblock URL \url{https://doi.org/10.1111/j.1467-8721.2008.00582.x}.

\bibitem[Peterson \& Peterson(1959)Peterson and Peterson]{trace-decay-peter}
Peterson, L.~R. and Peterson, M.~J.
\newblock Short-term retention of individual verbal items.
\newblock \emph{J Exp Psychol}, 58:\penalty0 193--198, September 1959.

\bibitem[Poo et~al.(2016)Poo, Pignatelli, Ryan, Tonegawa, Bonhoeffer, Martin, Rudenko, Tsai, Tsien, Fishell, Mullins, Gon{\c c}alves, Shtrahman, Johnston, Gage, Dan, Long, Buzs{\'a}ki, and Stevens]{what-is-memory-engram}
Poo, M.-m., Pignatelli, M., Ryan, T.~J., Tonegawa, S., Bonhoeffer, T., Martin, K.~C., Rudenko, A., Tsai, L.-H., Tsien, R.~W., Fishell, G., Mullins, C., Gon{\c c}alves, J.~T., Shtrahman, M., Johnston, S.~T., Gage, F.~H., Dan, Y., Long, J., Buzs{\'a}ki, G., and Stevens, C.
\newblock What is memory? the present state of the engram.
\newblock \emph{BMC Biology}, 14\penalty0 (1):\penalty0 40, 2016.
\newblock \doi{10.1186/s12915-016-0261-6}.
\newblock URL \url{https://doi.org/10.1186/s12915-016-0261-6}.

\bibitem[Raaijmakers \& Shiffrin(1980)Raaijmakers and Shiffrin]{sam}
Raaijmakers, J.~G. and Shiffrin, R.~M.
\newblock Sam: A theory of probabilistic search of associative memory.
\newblock volume~14 of \emph{Psychology of Learning and Motivation}, pp.\  207--262. Academic Press, 1980.
\newblock \doi{https://doi.org/10.1016/S0079-7421(08)60162-0}.
\newblock URL \url{https://www.sciencedirect.com/science/article/pii/S0079742108601620}.

\bibitem[Raaijmakers \& Shiffrin(1981)Raaijmakers and Shiffrin]{sam-origin}
Raaijmakers, J.~G. and Shiffrin, R.~M.
\newblock Search of associative memory.
\newblock \emph{Psychological Review}, 88\penalty0 (2):\penalty0 93--134, 1981.
\newblock \doi{10.1037/0033-295X.88.2.93}.
\newblock URL \url{https://doi.org/10.1037/0033-295X.88.2.93}.

\bibitem[Radford et~al.(2018)Radford, Narasimhan, Salimans, and Sutskever]{gpt1}
Radford, A., Narasimhan, K., Salimans, T., and Sutskever, I.
\newblock Improving language understanding by generative pre-training.
\newblock 2018.
\newblock URL \url{https://www.cs.ubc.ca/~amuham01/LING530/papers/radford2018improving.pdf}.

\bibitem[Rae et~al.(2016)Rae, Hunt, Danihelka, Harley, Senior, Wayne, Graves, and Lillicrap]{sparse-dnc}
Rae, J., Hunt, J.~J., Danihelka, I., Harley, T., Senior, A.~W., Wayne, G., Graves, A., and Lillicrap, T.
\newblock Scaling memory-augmented neural networks with sparse reads and writes.
\newblock In Lee, D., Sugiyama, M., Luxburg, U., Guyon, I., and Garnett, R. (eds.), \emph{Advances in Neural Information Processing Systems}, volume~29. Curran Associates, Inc., 2016.
\newblock URL \url{https://proceedings.neurips.cc/paper_files/paper/2016/file/3fab5890d8113d0b5a4178201dc842ad-Paper.pdf}.

\bibitem[Rae et~al.(2018)Rae, Dyer, Dayan, and Lillicrap]{fast-parametric}
Rae, J., Dyer, C., Dayan, P., and Lillicrap, T.
\newblock Fast parametric learning with activation memorization.
\newblock In Dy, J. and Krause, A. (eds.), \emph{Proceedings of the 35th International Conference on Machine Learning}, volume~80 of \emph{Proceedings of Machine Learning Research}, pp.\  4228--4237. PMLR, 10--15 Jul 2018.
\newblock URL \url{https://proceedings.mlr.press/v80/rae18a.html}.

\bibitem[Rae et~al.(2020)Rae, Potapenko, Jayakumar, Hillier, and Lillicrap]{compformer}
Rae, J.~W., Potapenko, A., Jayakumar, S.~M., Hillier, C., and Lillicrap, T.~P.
\newblock Compressive transformers for long-range sequence modelling.
\newblock In \emph{International Conference on Learning Representations}, 2020.
\newblock URL \url{https://openreview.net/forum?id=SylKikSYDH}.

\bibitem[Ramsauer et~al.(2021)Ramsauer, Sch{\"a}fl, Lehner, Seidl, Widrich, Gruber, Holzleitner, Adler, Kreil, Kopp, Klambauer, Brandstetter, and Hochreiter]{hopfield-all-you-need}
Ramsauer, H., Sch{\"a}fl, B., Lehner, J., Seidl, P., Widrich, M., Gruber, L., Holzleitner, M., Adler, T., Kreil, D., Kopp, M.~K., Klambauer, G., Brandstetter, J., and Hochreiter, S.
\newblock Hopfield networks is all you need.
\newblock In \emph{International Conference on Learning Representations}, 2021.
\newblock URL \url{https://openreview.net/forum?id=tL89RnzIiCd}.

\bibitem[Roediger(1978)]{retrieval-as-self-limiting-process}
Roediger, H.~L.
\newblock Recall as a self-limiting process.
\newblock \emph{Memory \& Cognition}, 6\penalty0 (1):\penalty0 54--63, 1978.
\newblock \doi{10.3758/BF03197428}.
\newblock URL \url{https://doi.org/10.3758/BF03197428}.

\bibitem[Roediger \& Butler(2011)Roediger and Butler]{role-of-retrieval-practice}
Roediger, H.~L. and Butler, A.~C.
\newblock The critical role of retrieval practice in long-term retention.
\newblock \emph{Trends in Cognitive Sciences}, 15\penalty0 (1):\penalty0 20--27, 2011.
\newblock ISSN 1364-6613.
\newblock \doi{https://doi.org/10.1016/j.tics.2010.09.003}.
\newblock URL \url{https://www.sciencedirect.com/science/article/pii/S1364661310002081}.

\bibitem[Rumelhart \& McClelland(1987)Rumelhart and McClelland]{rnn}
Rumelhart, D.~E. and McClelland, J.~L.
\newblock \emph{Learning Internal Representations by Error Propagation}, pp.\  318--362.
\newblock 1987.

\bibitem[Rundus \& Atkinson(1970)Rundus and Atkinson]{rehearsal-process-in-free-recall}
Rundus, D. and Atkinson, R.~C.
\newblock Rehearsal processes in free recall: A procedure for direct observation.
\newblock \emph{Journal of Verbal Learning and Verbal Behavior}, 9\penalty0 (1):\penalty0 99--105, 1970.
\newblock ISSN 0022-5371.
\newblock \doi{https://doi.org/10.1016/S0022-5371(70)80015-9}.
\newblock URL \url{https://www.sciencedirect.com/science/article/pii/S0022537170800159}.

\bibitem[Shiffrin \& Atkinson(1969)Shiffrin and Atkinson]{storage-and-retrieval}
Shiffrin, R.~M. and Atkinson, R.~C.
\newblock Storage and retrieval processes in long-term memory.
\newblock \emph{Psychological Review}, 76\penalty0 (2):\penalty0 179--193, 1969.
\newblock \doi{10.1037/h0027277}.
\newblock URL \url{https://doi.org/10.1037/h0027277}.

\bibitem[Shiffrin \& Raaijmakers(1992)Shiffrin and Raaijmakers]{sam-retro}
Shiffrin, R.~M. and Raaijmakers, J.
\newblock \emph{The SAM retrieval model: A retrospective and prospective.}, pp.\  69--86.
\newblock Essays in honor of William K. Estes, Vol. 1: From learning theory to connectionist theory; Vol. 2: From learning processes to cognitive processes. Lawrence Erlbaum Associates, Inc, Hillsdale, NJ, US, 1992.
\newblock ISBN 0-8058-1097-8 (Hardcover); 0-8058-0759-4 (Hardcover).

\bibitem[Song et~al.(2000)Song, Miller, and Abbott]{hebb-competitive}
Song, S., Miller, K.~D., and Abbott, L.~F.
\newblock Competitive hebbian learning through spike-timing-dependent synaptic plasticity.
\newblock \emph{Nature Neuroscience}, 3\penalty0 (9):\penalty0 919--926, 2000.
\newblock \doi{10.1038/78829}.
\newblock URL \url{https://doi.org/10.1038/78829}.

\bibitem[Underwood \& Postman(1960)Underwood and Postman]{interference}
Underwood, B.~J. and Postman, L.
\newblock Extraexperimental sources of interference in forgeting.
\newblock \emph{Psychol Rev}, 67:\penalty0 73--95, March 1960.

\bibitem[van~de Ven et~al.(2020)van~de Ven, Siegelmann, and Tolias]{brain-inspired-replay}
van~de Ven, G.~M., Siegelmann, H.~T., and Tolias, A.~S.
\newblock Brain-inspired replay for continual learning with artificial neural networks.
\newblock \emph{Nature Communications}, 11\penalty0 (1):\penalty0 4069, 2020.
\newblock \doi{10.1038/s41467-020-17866-2}.
\newblock URL \url{https://doi.org/10.1038/s41467-020-17866-2}.

\bibitem[Vaswani et~al.(2017)Vaswani, Shazeer, Parmar, Uszkoreit, Jones, Gomez, Kaiser, and Polosukhin]{transformer}
Vaswani, A., Shazeer, N., Parmar, N., Uszkoreit, J., Jones, L., Gomez, A.~N., Kaiser, L.~u., and Polosukhin, I.
\newblock Attention is all you need.
\newblock In Guyon, I., Luxburg, U.~V., Bengio, S., Wallach, H., Fergus, R., Vishwanathan, S., and Garnett, R. (eds.), \emph{Advances in Neural Information Processing Systems}, volume~30. Curran Associates, Inc., 2017.
\newblock URL \url{https://proceedings.neurips.cc/paper_files/paper/2017/file/3f5ee243547dee91fbd053c1c4a845aa-Paper.pdf}.

\bibitem[Waugh \& Norman(1965)Waugh and Norman]{displacement}
Waugh, N.~C. and Norman, D.~A.
\newblock Primary memory.
\newblock \emph{Psychological Review}, 72\penalty0 (2):\penalty0 89--104, 1965.
\newblock \doi{10.1037/h0021797}.

\bibitem[Wu et~al.(2022)Wu, Rabe, Hutchins, and Szegedy]{memorizing}
Wu, Y., Rabe, M.~N., Hutchins, D., and Szegedy, C.
\newblock Memorizing transformers.
\newblock In \emph{International Conference on Learning Representations}, 2022.
\newblock URL \url{https://openreview.net/forum?id=TrjbxzRcnf-}.

\bibitem[Zaheer et~al.(2020)Zaheer, Guruganesh, Dubey, Ainslie, Alberti, Ontanon, Pham, Ravula, Wang, Yang, and Ahmed]{bigbird}
Zaheer, M., Guruganesh, G., Dubey, K.~A., Ainslie, J., Alberti, C., Ontanon, S., Pham, P., Ravula, A., Wang, Q., Yang, L., and Ahmed, A.
\newblock Big bird: Transformers for longer sequences.
\newblock In Larochelle, H., Ranzato, M., Hadsell, R., Balcan, M., and Lin, H. (eds.), \emph{Advances in Neural Information Processing Systems}, volume~33, pp.\  17283--17297. Curran Associates, Inc., 2020.
\newblock URL \url{https://proceedings.neurips.cc/paper_files/paper/2020/file/c8512d142a2d849725f31a9a7a361ab9-Paper.pdf}.

\end{thebibliography}
